\documentclass[journal]{IEEEtran}
\usepackage{cite}
\usepackage{amsmath,amssymb,amsfonts}
\usepackage{booktabs} 
\usepackage{algorithm}
\usepackage{algorithmic}
\usepackage[pdftex]{graphicx}
\usepackage{textcomp}
\usepackage{xcolor}
\usepackage{url}

\begin{document}
%
\title{CNFP: Optimizing Cloud-Native Network Function Placement with Diffusion Models on the Cloud Continuum}
%
%
%

\author{Álvaro~Vázquez-Rodríguez,
        Manuel~Fernández-Veiga,
        Carlos~Giraldo-Rodríguez
\thanks{The work done is supported by the Axencia Galega de Innovación (GAIN) (11/IN606D/2023/2547030) and by grant PID2023-148716OB-C31 funded by MCIU/AEI/10.13039/501100011033 (project DISCOVERY).}
\thanks{Álvaro Vázquez-Rodríguez and Manuel Fernández-Veiga are with atlanTTic - I\&C Lab - Universidade de Vigo}%
\thanks{Álvaro Vázquez-Rodríguez and Carlos Giraldo-Rodríguez are with Centro Tecnolóxico de Telecomunicacións de Galicia (GRADIANT), Vigo, Spain}
}

%
%

%

\maketitle

\begin{abstract}
The placement of Cloud-Native Network Functions (CNFs) across the Cloud-Continuum represents a core challenge in the orchestration of current 5G and future 6G networks. The process entails the implementation of interdependent computing tasks, which are structured as Service Function Chains, over distributed cloud infrastructures. This is achieved while satisfying strict resource, bandwidth, connectivity, and end-to-end latency constraints. It is widely acknowledged that classical approaches, including mixed-integer (non)linear programming, heuristics, and reinforcement learning, face practical limitations in terms of scalability, robust constraint handling, and generalization to unseen network conditions. In this study, a diffusion-based theoretical and algorithmic framework for CNF placement is proposed, based on Denoising Diffusion Probabilistic Models (DDPM). The placement process is reconceptualised as a conditional graph-to-assignment generation task. Each scenario is encoded as a heterogeneous graph, capturing infrastructure and service-chain structure. A Graph Neural Network (GNN) denoiser is trained to iteratively refine noisy \textit{CNF-to-cloud} assignment matrices. In order to bias the generation process towards valid deployments, the model incorporates constraint-aware penalties during training. At inference, a multitude of candidate placements are sampled, and the best suboptimal, feasible solution is selected. This enables a controllable trade-off between solution quality and runtime. Extensive experimentation on diverse topologies, incorporating out-of-distribution evaluations with larger instances and shifted constraint regimes, demonstrates that the proposed approach consistently generates feasible solutions with considerably accelerated inference compared to MINLP solvers when available, while maintaining robust feasibility under constraint-tight scenarios. The findings of this study demonstrate the potential of diffusion-based generative modelling as a scalable tool for constrained network placement and embedding in cloud-continuum orchestration.
\end{abstract}


\IEEEpeerreviewmaketitle

\section{Introduction}
\label{sec:intro}

\IEEEPARstart{N}{etwork} operators are undergoing a significant transformation in the deployment of 5G networks. This transformation is characterised by the adoption of software-centric principles and a shift towards cloud-native design methodologies. The transition is marked by a departure from the utilisation of monolithic appliances and VM-based implementations, towards a microservice-oriented paradigm. This shift is exemplified by the adoption of Network Function Virtualisation (NFV) and containerised network functions, which are designed to enhance operational efficiency and flexibility. In this context, it is becoming increasingly commonplace for 5G system functions to be implemented as software network functions. These functions can be instantiated and scaled on demand, with coordination facilitated by NFV Management and Orchestration (NFV-MANO) frameworks~\cite{etsi_nfv_mano,3gpp23501}. Concurrently, the transition towards Multi-access Edge Computing (MEC) and fog/edge paradigms facilitates the migration of computation from central clouds to resource-constrained nodes in closer proximity to end users. This development enables the provision of latency-sensitive services at the network edge~ \cite{etsi_mec003,bonomi2012fog}. This evolution gives rise to a heterogeneous cloud continuum, in which functions may be distributed across both core and edge sites, with the consequence that these sites are subject to diverse resource, connectivity, and operational cost constraints.

In order to deliver end-to-end services, network functions are frequently composed into Service Function Chains (SFCs). In such cases, traffic must traverse an ordered (or partially ordered) set of functions. The collective provision of these functions results in a higher-level network service~\cite{bhamare2016sfc}. Determining the location of required functions across the cloud continuum, whilst satisfying constraints such as node capacity (e.g. CPU/RAM), link bandwidth, and stringent end-to-end delay budgets, gives rise to challenging combinatorial optimisation problems. These problems are widely recognised as difficult and are commonly addressed through mixed-integer optimisation, heuristics/metaheuristics, or learning-based approximations~\cite{laghrissi2019vnfplacement,bhamare2016sfc}..
Exact mixed-integer (and mixed-integer nonlinear) formulations have been shown to provide optimal solutions for small instances. However, as the number of nodes, functions and constraints increases, these formulations scale poorly. This motivates the use of faster heuristics that are tailored to specific scenarios~\cite{laghrissi2019vnfplacement}. In recent times, the application of reinforcement learning to cope with dynamism and large state spaces has been explored. However, this typically requires carefully designed reward functions and still struggles to enforce strict feasibility under hard constraints~\cite{santos2020rlsfc}.

Recently, solutions to these problems based on diffusion models have begun to emerge 
as a competitive alternative. Diffusion probabilistic models are a type of machine learning model that learns a data distribution by reversing a gradual noising process. During the training phase, noise is progressively injected into a data sample, while a neural denoiser is optimised to predict and remove that noise. At the inference phase, generation starts from noise and iteratively denoises to synthesise new samples~\cite{ddpm}. This iterative generation is attractive for structured decision-making because it can represent multi-modal solution spaces and, through parallel sampling, produce diverse candidates from which the best feasible solution can be selected under system constraints. Beyond
the original DDPM formulation, score-based perspectives unify diffusion and enable
alternative samplers and guidance mechanisms that can trade off sample quality and
runtime~\cite{song2021scorebased}. Recent research has demonstrated the efficacy of graph-conditioned diffusion in solving NP-hard combinatorial problems. This is achieved by learning to denoise structured solutions on graphs and improving quality via sampling and selection~\cite{difusco}. The concept of diffusion as a general solution generator for network optimisation has been proposed in the context of networking. The model in question learns a conditional solution distribution, with sampling concentrating probability mass around high-quality decisions~\cite{liang2025diffsg,liangexplorations}. It is important to note that graph diffusion models are capable of being trained from suboptimal or heuristic datasets, with the result that near-optimal solutions are recovered. This, in turn, has the effect of reducing the reliance on expensive exact solvers~\cite{GDSG,hdu}. The most closely related work to the problem under discussion is that of Network Diffuser, which applies conditional diffusion to the placement and scheduling of service function chains using inverse demonstration. This highlights the suitability of diffusion for SFC decision generation under constraints ~\cite{zhang}. The motivation behind the present study was provided by the results of earlier research. The latter constituted an examination of the cloud continuum, with the objective of determining the feasibility of casting static CNF/SFC placement in this setting as a conditional generation problem. This was achieved by means of a graph-based denoiser, which learned to map infrastructure and service-chain structure to a distribution over placement assignments. The result of this learning process was the enabling of fast inference through sampling and post-selection.

Based on these ideas, the present article proposes a solver based on diffusion models for SFC/CNF placement across the cloud continuum. Specifically, the contributions can be categorised into three distinct aspects. Firstly, a static placement problem for multiple SFCs over a cloud-continuum infrastructure graph is formalised, capturing heterogeneous node capacities and costs, CNF resource demands and processing delays, placement restrictions, and end-to-end feasibility requirements under CPU/RAM, adjacency, bandwidth, and SFC delay constraints. Secondly, a conditional diffusion-based generative solver is proposed, with each instance represented as a heterogeneous graph (cloud nodes, CNF nodes, and SFC edges) and capable of learning to denoise a CNF-to-cloud assignment matrix. A graph neural network (GNN) encoder processes the infrastructure and service-chain structure, while a placement mask enforces forbidden CNF-cloud pairs and additional constraint-aware loss terms bias training towards feasible, high-quality solutions. At inference, multiple samples are generated and the best feasible placement is selected. Thirdly, an extensive experimental investigation was conducted on 44 scenarios. A GEKKO-based MINLP solver acted as the optimal reference when available, and this was compared against two greedy heuristics and a range of learning-based baselines. To assess scalability and generalisation in cost- and constraint-shifted regimes, out-of-distribution scenarios were employed.

The remainder of the paper is organized as follows. First, in 
Section~\ref{sec:related-work}, we present the main prior works related to our
contributions. How we model the network of our system, how CNFs and SFCs are defined 
and the formulation of the problem are presented in Section~\ref{sec:system-model}. 
The design and implementation of the solver based on diffusion models and Graph 
Neural Networks (GNNs) is explained in Section~\ref{sec:diffusion-solver}, and the
explanation of how we evaluate, compare and obtain results to measure our model can 
be found in Section~\ref{sec:results}. Concluding remarks, actual limitations and future work are given in Section~\ref{sec:conclusions}. 

\section{Related work}
\label{sec:related-work}

The intersection of diffusion models approaches and network optimisation has expanded
rapidly in the past years, fuelled by the success of Denoising Diffusion Probabilistic
Models (DDPMs)~\cite{ddpm} in combinatorial tasks. Surveys on virtual network embedding 
and service function chain deployment already note a shift from classical heuristics 
toward learning-based solvers~\cite{satpathy}, and several works now explore diffusion
techniques specifically for resource-allocation and routing
problems~\cite{liangexplorations}. 

One of the earliest framework in this path is~\cite{liang2025diffsg}. DiffSG trains 
a continuous-time DDPM conditioned on an optimisation instance so that low cost 
solutions become high probability points in the learned distribution. At inference, 
the model generates multiple candidate vectors by iterative denoising and selects the 
best, consistently surpassing mixed-integer programming and heuristic baselines on 
diverse benchmarks. Because training only requires a scalar cost evaluator, DiffSG 
is agnostic to the exact form of the objective and naturally accommodates multi 
objective or non-linear formulations, an attractive property for CNF/SFC placement 
where latency, cost and energy often interplay. 

A complementary line of research embeds diffusion models inside reinforcement learning 
(RL) pipelines. Deep Diffusion Soft Actor–Critic (D2SAC)~\cite{hdu} augments the actor 
of Soft Actor–Critic with a DDPM that proposes multimodal action sequences. The richer
policy distribution accelerates convergence and improves final rewards on standard control
benchmarks, implying that similar hybrids could enhance RL-based network orchestration 
by sampling diverse placement schedules that simple Gaussian policies cannot represent. 

Because many network embedding problems are graph structured, several studies pair
diffusion processes with GNNs. GDSG~\cite{GDSG} tackles multi-hop computation 
off-loading by placing a GNN-based denoiser atop a Gaussian diffusion chain; trained 
solely on heuristic solutions, the model recovers near-optimal task assignments with 
high probability. DIFUSCO~\cite{difusco} generalises this idea to canonical NP-hard
problems such as the Travelling Salesman Problem and Maximum Independent Set, cutting 
the optimality gap of previous neural solvers by roughly 50\% on graphs with thousands 
of nodes. These successes indicate that graph conditioned diffusion can efficiently 
explore huge discrete spaces, a characteristic shared with CNF placement.

\begin{table*}[t]
\centering
\caption{Comparison with state of the art solutions}
\label{tab:rw-contrast}
\footnotesize
\begin{tabular}{p{2cm}p{2.5cm}p{2.5cm}p{4cm}p{4cm}}
\toprule
\textbf{Work} & \textbf{Problem focus} & \textbf{Output type} & \textbf{Conditioning / data} & \textbf{Constraint handling} \\
\midrule
\textbf{DiffSG}~\cite{liang2025diffsg} & Generic network optimization & Continuous vectors, with discretization for binary & Black-box cost evaluator; instance-conditioned diffusion; no optimal labels needed & Constraints folded into objective via penalties; sampling selects low-cost solutions \\ 
\textbf{GDSG}~\cite{GDSG} & MEC multi-hop computation offloading: discrete offload + continuous alloc. & Hybrid discrete + continuous heads & GNN-conditioned diffusion; suboptimal heuristic data suffices; task orthogonality between heads & Feasibility promoted via training loss design; final selection over samples \\ 
\textbf{Network Diffuser}~\cite{zhang} & Online SFC placing \& scheduling & Trajectory/state sequences & Conditional diffusion over state rollouts; inverse demonstrations & Feasibility via conditioning and guidance during denoising \\
\textbf{DIFUSCO}~\cite{difusco} & Classic CO on graphs (TSP, MIS, etc.) & Discrete (binary/perm.) & Graph diffusion (Gaussian/Bernoulli) on node/edge vars & Validity encouraged by process design + schedules \\ 
\textbf{This work} & CNF/SFC placement on Cloud–Continuum with restrictions & Discrete CNF$\to$cloud assignment & Heterogeneous graph (cloud+CNF) conditions a DDPM; no labels (noise MSE) + constraint losses; best-of-$k$ & Hard constraints enforced by mask + penalties; choose best feasible sample \\
\bottomrule
\end{tabular}
\end{table*}

Beyond diffusion, learning-based methods continue to advance state of the art 
placement in novel environments. For Low Earth Orbit (LEO) satellite constellations, 
Doan et al.~\cite{Doan} combine multi-agent Q-learning for chain scheduling with 
Bayesian optimisation for Virtual Network Function (VNF) caching. Although diffusion 
models have not yet been applied in LEO scenarios, their ability to sample global,
constraint aware configurations suggests a promising future direction. 

Very recently, Network Diffuser~\cite{zhang} applies conditional diffusion directly to
online SFC placement and scheduling. The authors view the problem as generating a 
feasible state trajectory: a DDPM denoises random sequences of placements into valid 
ones while conditioning on resource limits and latency requirements. Lacking expert 
traces, they create synthetic training data via inverse optimisation. It demonstrates 
the practicality of tailoring diffusion to complex SFC placement tasks.

In summary, existing work shows that diffusion models can learn expressive, 
conditioned distributions over feasible solutions, yielding sub-optimal performance 
across routing, offloading and scheduling domains. Nevertheless, no prior study has
addressed CNF placement across the Cloud-Continuum. The present paper bridges this gap 
by designing a topology-aware, GNN-based DDPM that generates full CNF mappings on the 
clouds available. Table~\ref{tab:rw-contrast} contrasts diffusion-based optimizers 
most relevant to our study with classical VNE/SFC lines. We differ from previous works
by targeting CNF placement across the Cloud–Continuum with strict constraints, solved via a graph-conditioned DDPM that outputs full CNF$\to$cloud assignments and is evaluated against an exact MINLP under a fixed time budget.

\section{System Model}
\label{sec:system-model}

\begin{figure}[b]
\centering
\includegraphics[width=0.5\columnwidth]{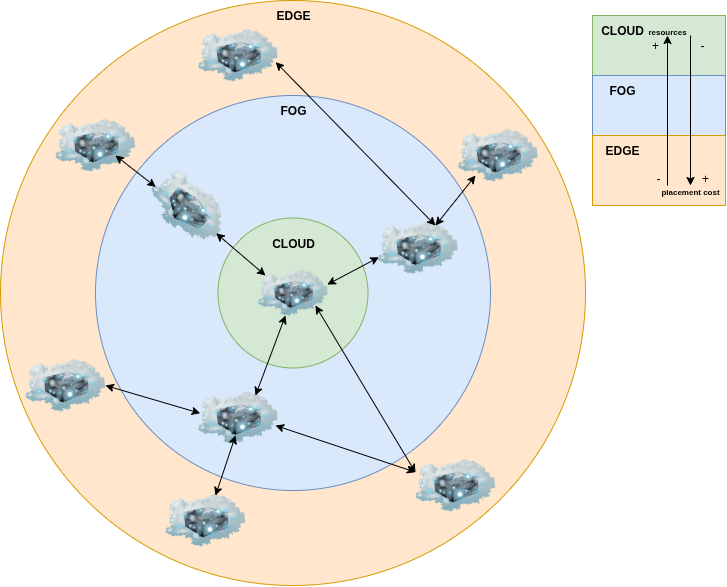}
\caption{Network model of the problem.}\label{fig:infrastructure}
\end{figure}

\subsection{Network model}
The system model described in Figure~\ref{fig:infrastructure} and formalized in Table~\ref{tab:notation} is used for formalizing 
the problem of CNF placement under constraints. We assume that the system consists 
of multiple clouds $\mathcal{C} = \{ 1, 2, \dots, C \}$ where the computational tasks 
of an atomic CNF are placed, instantiated and executed. The clouds in $\mathcal{C}$ 
are independently managed and operated, but we assume that they are connected in an
arbitrary graph $\mathcal{G}_\mathcal{C}$ such that at least one path exists from 
cloud $i$ to cloud $j$, for every $i, j \in \mathcal{C}$. The set of neighbours of 
cloud $i$ is denoted as $\mathcal{N}_i$, for any $i \in \mathcal{C}$. We model the
connections between two clouds, for simplicity, as a single link with capacity 
$c_{ij}$ bits/s.

The graph $\mathcal{G}_\mathcal{C}$ is static and predetermined, i.e., it is assumed 
that the communication network among the clouds does not change over the lifetime of 
the system. This corresponds to a static view of the scheduling and placement problem, 
in which the long-term performance goals have a well-defined steady state that, 
when reached, does no change with time.

\subsection{SFCs and CNFs}
\begin{figure}[b!]
\centering
\includegraphics[width=0.75\columnwidth]{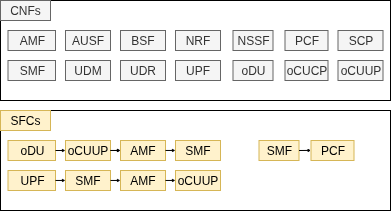}
\caption{Example of CNFs and SFCs}\label{fig:cnfsfc}
\end{figure}

For the purposes of this work, a SFC $s_h$, $h = 1, \dots, H$, is a partially 
ordered sequence of atomic computing tasks, i.e., each constituent computing task cannot 
be split or divided further into smaller, separable tasks. This individual, 
non-decomposable tasks are called CNFs, and we assume without loss of generality that 
the CNFs can be taken from a finite set of $M$ possible CNFs, $\mathcal{F} = \{ f_1, 
\dots, f_M \}$. An example of CNFs and SFCs can be seen in Fig.~\ref{fig:cnfsfc}

An SFC $s_h$ is modelled as a directed acyclic graph (DAG) $\mathcal{S}_h$ with one
starting node and one terminating node. In other words, $\mathcal{S}_h$ has a root node, 
a sink node, and a number of intermediate nodes, with the condition that there are no
cycles in $\mathcal{S}_h$. In the DAG $\mathcal{S}_h$, node $u \in \mathcal{F}$ precedes
node $v \in \mathcal{F}$ if there is an edge $u \rightarrow v$. The DAG has a natural
\emph{topological ordering} of its nodes, $\mathsf{ord}(\mathcal{S}_h) = (s_h^{(1)},
s_h^{(2)}, \dots, s_h^{(\ell_h)})$, wherein if $u \rightarrow v$, then $u$ appears
before $v$ in the list $\mathsf{ord}(\mathcal{S}_h)$.

We further assume that each individual CNF $s_h^{(j)}$ in a SFC $\mathcal{S}_h$ can 
be placed in a cloud from $\mathcal{C}$, but needs some amount of computing resources
$\mathbf{z}_h^{(j)} = (x_h^{(j)}, d_h^{(j)}, \mathbf{r}_h^{(j)})$ where $x_h^{(j)}$ 
is the amount of computing resource (e.g. number of CPUs), $d_h^{(j)}$ is the amount 
of storage units (e.g. amount of RAM), and $\mathbf{r}_h^{(j)}$ is the vector of 
communication rates from CNF $s_h^{(j)}$ to each of its subsequent CNFs in the SFC. 
To account for possible restrictions on the usage of resources, we also assume that 
a cloud $i \in \mathcal{C}$ can only execute a predefined set of CNF types 
$\mathcal{T}_i \subseteq \mathcal{F}$. Thus, a CNF $s_h^{(j)}$ can only be placed at 
cloud $i$ if $s_h^{(j)} \in \mathcal{T}_i$. Obviously, the case $\mathcal{T}_i =
\mathcal{F}$ means that any cloud can host an arbitrary CNF of any type. Additionally,
observe that in our model the resource vector $\mathbf{z}_h^{(j)}$ is only dependent 
on the type of the CNF, regardless its placement in the available clouds.

In our system model, a fixed collection of SFCs $\mathcal{S} = \{ s_1, s_2, \dots, s_H \}$
is known in advance. This collection must be placed in the system without violating the
resource constraints, the partial topological ordering of the constituent CNFs for each
SFC, while optimising a cost function described below.

\begin{table}[b]
\caption{System model notation}
\label{tab:notation}
\centering
\small
\begin{tabular}{ll}
\toprule
Symbol & Meaning \\
\midrule
$\mathcal{C}$, $C$ & Set / number of clouds \\
$\mathcal{F}$, $M$ & Set / number of CNFs \\
$\mathcal{S}$, $H$ & Set / number of service function chains \\
$X_i, D_i$ & CPU / RAM capacity of cloud $i$ \\
$x_m, d_m$ & CPU / RAM demand of CNF $m$ \\
$t_{i,m}$ & Cost of placing CNF $m$ at cloud $i$ \\
$c_{i,j}$ & Bandwidth capacity on link $(i,j)$ \\
$\tau_m$ & Processing delay of CNF $m$ \\
$r_{u,v}$ & Rate demand for hop $(u\!\to\!v)$ \\
$\tau_h$ & End-to-end delay budget of chain $h$ \\
\bottomrule
\end{tabular}
\end{table}

\subsection{Problem formulation}
\label{sec:problem-formulation}

\begin{figure}[t]
\centering
\includegraphics[width=0.75\columnwidth]{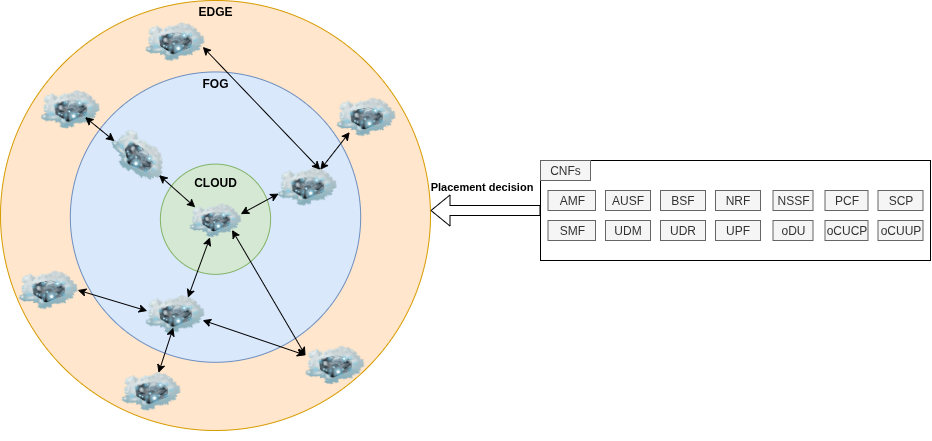}
\caption{CNF placement problem on the Cloud-Continuum}\label{fig:placement}
\end{figure}

We will assume that the placement of a CNF $s_h^{(j)}$ at cloud $i$ has a fixed 
cost $t_{im}$, where $m = s_h^{(j)}$ is the type of the CNF. The execution of a 
computing task at the CNF has a fixed delay $\tau_m$, and the delay for transmitting 
a message from a CNF running in cloud $i$ to a CNF running in cloud $j$ is equal to 
$\kappa / c_{ij}$, where $\kappa$ is a suitable but arbitrary constant that can be
interpreted as the size (in bits) of the message sent from $i$ to $j$, supposed of 
constant length here.

Let $x_{ijh} = 1$ if CNF $s_h^{(j)}$ is placed at cloud $i$, and $x_{ijh} = 0$ 
otherwise. The problem of optimal placement (Fig.~\ref{fig:placement}) of the collection 
of SFCs $\mathcal{S}$ on the cloud federation $\mathcal{C}$ with static topology
$\mathcal{G}_\mathcal{C}$ can be formulated as follows
\begin{IEEEeqnarray}{l}
 \min \sum_{h = 1}^H x_{ijh} t_{im} \IEEEyesnumber \label{sc-placement:cost} \\ \IEEEyessubnumber*
 \sum_{i \in C} x_{ijh} = 1, \qquad \forall\, h \in [H],\; j \in [\ell_h] \label{sc-placement:one} \\ 
 x_{ijh} e_{jj^\prime}^{(h)} x_{i^\prime j^\prime h} \leq e_{ii^\prime},  \forall i, i^\prime \in \mathcal{C}, h \in [H], j \neq j^\prime \in [\ell_h] \label{sc-placement:dag} \\
 \sum_{h = 1}^H x_{ijh} x_h^{(j)} \leq X_i, \quad \forall i \in \mathcal{C} \label{sc-placement:resource1} \\
 \sum_{h = 1}^H x_{ijh} d_h^{(j)} \leq D_i, \quad \forall i \in \mathcal{C} \label{sc-placement:resource2} \\
 \sum_{h = 1}^H x_{ijh} e_{j j^\prime}^{(h)} x_{i^\prime j^\prime h} r_h^{(j j^\prime)} \leq c_{i i^\prime}, \ \forall i \in \mathcal{C}, j, j^\prime \in [\ell_h] \label{sc-placement:resource3} \\
 J(\mathbf{x}_h, \mathcal{G}_\mathcal{C}) \leq \tau_h, \ \forall h \in [H] \label{sc-placement:delay} \\
 J(\mathbf{x}_h, \mathcal{G}_\mathcal{C}) = \max_{p \in \mathcal{P}_h} \{ \sum_{i = 1}^C \sum_{q = 1}^{h_p - 1} (\tau_{i j_q}  +  \frac{\kappa}{c_{j_q j_{q+1}}} ) x_{i j_q h} \nonumber \\ 
 + \tau_{i j_{h_p}} x_{ij_{h_p} h} \} \label{sc-placement:delay-2} \\
 x_{ijh} = 0, \qquad \forall\, i \in \mathcal{C},\; \forall\, h \in [H],\; \nonumber \\
 \forall\, j \in [\ell_h]\ \text{s.t.}\ s_h^{(j)} \notin \mathcal{T}_i \label{sc-placement:restr} \\
 x_{ijh} \in \{ 0, 1 \} \ \forall i \in \mathcal{C}, h \in [H], j \in [\ell_h] \label{sc-placement:binary}
\end{IEEEeqnarray}
These equations have the following interpretation:
\begin{itemize}
\item \eqref{sc-placement:cost} is the cost function of the placement, extended to all 
the SFCs.

\item Constraints~\eqref{sc-placement:one} enforces that each CNF $s_h^{(j)}$ is placed 
in one and only one cloud $i$.

\item Constraints~\eqref{sc-placement:dag} are the topological restrictions: the CNFs 
from a  SFC $s_h$ must be placed such that a physical link exists between the placement $i$ 
of CNF $s_h^{(j)}$ and the placement $i^\prime$ of CNF $s_h^{(j^\prime)}$ if $s_h^{(j)}$
precedes $s_h^{(j^\prime)}$. The variables $e_{i i^\prime}$ are binary and equal $1$ if 
and only if the clouds $i$ and $i^\prime$ are directly connected. Similarly, the 
variables $e_{j j^\prime}^{(h)}$ are equal to $1$ if and only if $s_h^{(j)}$ precedes
$s_{h}^{(j^\prime)}$. Consecutive CNFs may be co-located in the same cloud, i.e., 
we assume \(e_{ii}=1\) so that zero-hop “links” are allowed.
  
\item Constraints~\eqref{sc-placement:resource1} and \eqref{sc-placement:resource2} are 
the constraints on the physical resources at cloud $i$.

\item \eqref{sc-placement:resource3} are the constraints on the communication rate 
between neighbouring clouds $i, i^\prime$: the sum of requested rates from all the 
CNFs placed at $i$ towards the CNFs placed at $i^\prime$ cannot exceed the available
bandwidth $c_{ii^\prime}$. 

\item \eqref{sc-placement:delay} are the delay constraints, where $\tau_h$ is the 
maximum allowed delay for completing SFC $s_h$. Here,~\eqref{sc-placement:delay-2} 
denotes the delay for the placement $\mathbf{x}_h := (x_{ijh}: i \in \mathcal{C}, j 
\in [\ell_h])$ on $\mathcal{G}_\mathcal{C}$. On the DAG that defines the SFC $s_h$ and 
the placement $\mathbf{x}_h$, this delay can be computed with the following expression: 
let $p = (j_1, j_2, \dots, j_{h_p})$ be a path with $h_p$ nodes through the DAG of the 
SFC $s_h$, where every index $j_q \in [\ell_h]$, $i_1$ is the root node and $j_{h_p}$ 
is the sink node. Then, the delay for completing this SC is the maximum delay along 
any path for SCF, given the placement $\mathbf{x}_h$.

\item Constraint~\eqref{sc-placement:restr} encodes CNF-to-cloud compatibility: a CNF of type $s_h^{(j)}$ may be placed at cloud $i$ only if $s_h^{(j)} \in \mathcal{T}_i$. The unconstrained case corresponds to $\mathcal{T}_i=\mathcal{F}$.
  
\item \eqref{sc-placement:binary} simply states that the placement decision
  variable is binary.
\end{itemize}

There exist some considerations to be taken into account on the above model:
\begin{enumerate}
\item We have assumed that CNFs are unique but a single CNF placed at a cloud is 
not dedicated to a single service function chain, i.e., a particular CNF can be 
shared by multiple SFCs or multiple requests.

\item In our mathematical model, we have not allowed the possibility that a cloud 
forwards a message to another cloud hosting the next CNF to be executed. In other words, 
we require that, if for a SFC $s_h$, the CNF $s_h^{(j)}$ precedes $s_h^{(j^\prime)}$, 
and they are placed at clouds $i$ and $i^\prime$, respectively, then there is a direct 
link between clouds $i$ and $i^\prime$.

\item Observe that the constraints~\eqref{sc-placement:dag} 
and~\eqref{sc-placement:resource3} can be linearized, but the delay 
function~\eqref{sc-placement:delay} is non-linear. The only case where it is linear is 
when the DAG is actually a line, i.e., the SFC is a sequential chain of CNFs.
\end{enumerate}

\section{Diffusion-based solver}
\label{sec:diffusion-solver}

\begin{figure}[t]
\centering
\includegraphics[width=0.75\columnwidth]{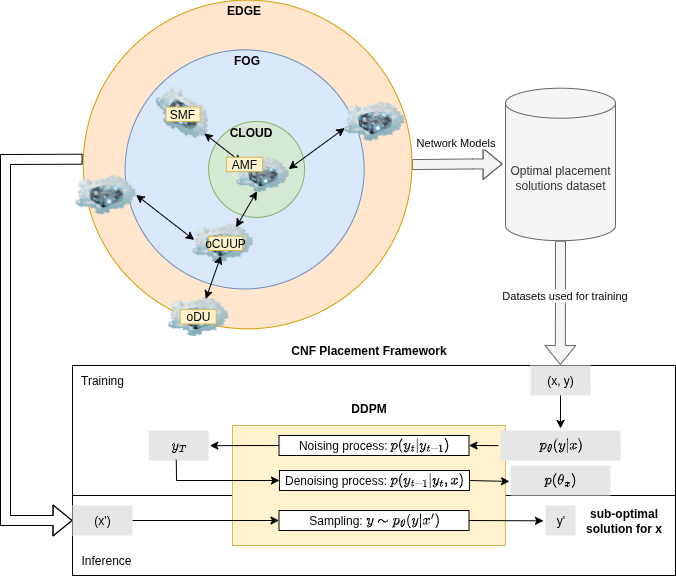}
\caption{Diffusion-based CNF placement framework architecture}\label{fig:modelo}
\end{figure}

In this Section, we present a diffusion-based framework for the CNF placement problem 
(see Figure~\ref{fig:modelo}). Unlike classical integer-programming solvers, our 
approach learns to generate feasible low-cost assignments by treating placement as a
generative task. We adapt the DDPM framework to this discrete combinatorial setting 
by using a GNN to encode the network topology, the CNF structure and by training 
with additional loss terms that enforce placement constraints.

Specifically, we design a neural network architecture that takes as input a heterogeneous
graph representing the network and CNFs, along with a assignment matrix. The model embeds 
node features via message-passing, incorporates the diffusion timestep, and outputs 
per-(CNF, cloud) scores interpreted as predicted noise. During training, we add 
Gaussian noise to the CNF-to-cloud assignment matrix and minimize the MSE between 
the model’s predicted noise and the true noise, following the DDPM paradigm. Crucially, 
we also integrate specific loss terms: penalties for capacity violations, adjacency
violations, bandwidth overuse, SFC delay budget violations and placement-restriction
violations, following the equations presented in Sec.~\ref{sec:problem-formulation}. 
At inference time, we run the reverse diffusion process to sample many candidate
assignments and pick the best valid solution.

\subsection{Graph-based Diffusion Architecture}

\subsubsection{Graph Representation}
To apply a diffusion model, we first represent the CNF placement instance as a graph
structured input. We use a heterogeneous graph $H$ with two node types: cloud nodes and 
CNF nodes.
\begin{itemize}
\item \emph{Cloud nodes}: Each cloud $i$ has a feature vector $[X_i, D_i, t_i]$ for 
its CPU, RAM and placement cost. We maintain a tensor of cloud type indices which is
embedded into a learned type embedding. Cloud–Cloud edges encode the network connectivity:
if bandwidth $c_{ij}>0$, we include a directed edge $(i\to j)$ with attribute $[c_{ij}]$.

\item \emph{CNF nodes}: Each CNF $f$ has a feature vector $[x_f, d_f, \tau_m]$ (CPU, 
RAM, processing delay). We have a restriction cloud placement index embedded into a 
learned vector. Within each SFC, we add edges between consecutive CNFs: for chain edges
$s_h^{j}\to s_h^{j+1}$, we include an edge with attributes $[\ell_{j,j+1}, 
\Delta_{s_h}, h, k]$ meaning the communication rate between consecutive CNFs, the SFC
delay, an SFC identifier and the hop index in the SFC.
\end{itemize}

Finally, we build a complete bipartite graph from each CNF to each allowed cloud node 
respecting placement restrictions. This heterogeneous graph captures all data: node features encode capacities and requirements; and edges encode topology: physical links 
and service chains.

\subsubsection{GNN Architecture}
The backbone model for propagating and aggregating information over $H$ is a GNN that 
predicts the added noise on the assignment matrix. It implements a conditional DDPM 
where the condition is the graph $H$. We include learnable embeddings for cloud type 
and for CNF restriction, and these are concatenated with the raw node features so that 
the model knows node types and any CNF placement restriction. Also, the diffusion 
timestep or noise level $\log\sigma_t$ is mapped through a two-layer multilayer 
perceptron (MLP) to produce a hidden embedding $t_{\text{emb}}$. This is then added to
every node’s embedding in each layer, allowing the GNN to be explicitly aware of the 
noise scale.

Two separate graph encoders process the cloud and CNF subgraphs. For clouds, we have 
an encoder with multiple GraphSAGE convolution layers augmented with edge attributes.
Similarly, a encoder for CNF with GraphSAGE layers processes the chain subgraph. 
Formally, let $x_c$ and $x_t$ be the augmented cloud and CNF feature matrices. 
The cloud encoder performs
\begin{equation}
    h_c = \mathrm{CloudEncoder}(x_c, E_{\text{cc}}, A_{\text{cc}}) \,
\end{equation}
where $E_{\text{cc}}$ and $A_{\text{cc}}$ are the cloud-cloud edge indices and bandwidth 
attributes. Likewise, the CNF encoder outputs
\begin{equation}
    h_t = \mathrm{CNFEncoder}(x_t, E_{\text{tt}}, A_{\text{tt}})
\end{equation}
for CNF–CNF edges. Each of these embeddings has dimension $H$. After encoding, we add 
the same time embedding $t_{\text{emb}}$ to all node vectors.

The network must reason about which CNF can go to which cloud. To mix information 
between cloud and CNF embeddings, we apply two GraphSAGE layers on the bipartite 
assignment edges. Let $E_{\text{tc}}$ denote the edge index of $CNF \rightarrow cloud$ 
assignment edges and $E_{\text{ct}}$ the reverse. We compute
\begin{IEEEeqnarray}{rCl}
    \IEEEyesnumber\IEEEyessubnumber*
    h_c' & = & h_c + \mathrm{SAGEConv}_{\text{t→c}}(h_t, h_c; E_{\text{tc}})\,,\quad \\
    h_t' & = & h_t + \mathrm{SAGEConv}_{\text{c→t}}(h_c', h_t; E_{\text{ct}})\,,
\end{IEEEeqnarray}
The model ultimately needs to predict a score for each possible (CNF, cloud) pair. 
To do this, we form the Cartesian product of the updated embeddings $h_t'$ and $h_c'$.
Concretely, for $M$ CNFs and $C$ clouds, we create $M\times C$ pairs $(h_{t_i}', h_{c_j}')$
and also include the current noisy assignment value $y_{i,j}$ as input. These vectors are
fed through a decoder MLP with two hidden layers and an activation function, producing a scalar
\textit{score} for each pair. The output is reshaped to an $M\times C$ matrix $S$. 
Thus, the model outputs unnormalized scores $s_{i,m}$ for assigning CNF $m$ to cloud $i$.

\subsection{Training process}
\label{sec:Training}

\subsubsection{Forward diffusion}

We use a standard Gaussian diffusion schedule with $T$ timesteps and precompute a 
cosine schedule $\sqrt{\bar\alpha_t}$ and $\sigma_t$ for $t=1,\dots,T$. The forward
noising process adds scaled Gaussian noise to the true assignment matrix 
$Y_0 \in \{0,1\}^{M \times C}$. At a random timestep $t$, we draw noise $E \sim 
\mathcal{N}(0,I)$, masked to valid placements, and set
\begin{equation}
    Y_t = \sqrt{\bar\alpha_t}\,Y_0 + \sigma_t\,E, 
\end{equation}
Then, it is applied the placement mask, where forbidden placements stay zero. This 
ensures $Y_t$ is a noisy real-valued matrix with the same sparsity pattern. The 
training goal is for the model to predict the noise $E$.

\subsubsection{Train}
The diffusion model is trained to minimize the expected squared error between the 
predicted noise and the true noise. Concretely, the network outputs a real-valued 
noise prediction $\hat{\epsilon}$ of the same shape as the noisy assignment. The denoising objective follows the standard $\epsilon$-prediction parametrization of DDPMs.
Given a random timestep $t$, we diffuse the assignment $Y_0$ to obtain
$Y_t$ and the injected Gaussian noise $E$. The GNN predicts $\widehat{E}=\epsilon_\theta(H,Y_t,\sigma_t)$,
and we minimize the mean-squared error
\begin{equation}
\mathcal{L}_{\text{denoise}}=\mathbb{E}_t\left[\left\|\widehat{E}-E\right\|_2^2\right].
\end{equation}

From the predicted noise, we reconstruct the clean assignment estimate
\begin{equation}
    \hat{Y_0} = (Y_t - \sigma_t\hat{\epsilon})/\sqrt{\bar{\alpha_t}}.
\end{equation} 
Then, we forbid illegal placements and obtain the probability matrix that feeds every constraint loss
\begin{equation}
    P = \operatorname{Softmax}(\hat{Y_0} \odot M - \infty\cdot(1-M))
\end{equation}
where $M$ is the placement mask. The full training procedure is summarised in Algorithm~\ref{alg:train}

\subsubsection{Constraint loss terms}
After predicting the noise $\widehat{E}$, we reconstruct
\begin{equation}
\widehat{Y}_0=\frac{Y_t-\sigma_t\widehat{E}}{\sqrt{\bar{\alpha}_t}},
\end{equation}
apply the placement mask (forbidden CNF-cloud pairs), and obtain a relaxed assignment matrix
$P\in[0,1]^{M\times C}$ via a row-wise softmax:
\begin{equation}
P=\text{softmax}(\text{Mask}(\widehat{Y}_0)),\qquad \sum_{i=1}^{C}P_{m,i}=1,\ \forall m.
\end{equation}
All constraint penalties are computed on $P$ and added to the denoising objective.

\medskip
\noindent\textbf{Placement restrictions.}
Let $\text{mask}_{m,i}\in\{0,1\}$ indicate whether CNF $m$ is allowed on cloud $i$.
Although masking is enforced before the softmax, we also penalize any residual probability mass assigned to forbidden pairs:
\begin{equation}
\mathcal{L}_{\text{restr}}(P)=\sum_{m=1}^{M}\sum_{i=1}^{C}\Big((1-\text{mask}_{m,i})\,P_{m,i}\Big)^{p},
\end{equation}

\medskip
\noindent\textbf{CPU/RAM capacity.}
Let $x_m$ and $d_m$ be CPU/RAM demands of CNF $m$, and $X_i$, $D_i$ the CPU/RAM capacities of cloud $i$.
The expected resource usage under $P$ is
\begin{equation}
U^{\text{cpu}}_i(P)=\sum_{m=1}^{M}P_{m,i}\,x_m,\qquad
U^{\text{ram}}_i(P)=\sum_{m=1}^{M}P_{m,i}\,d_m
\end{equation}

We penalize capacity violation with a mean ReLU:
\begin{multline}
\mathcal{L}_{\text{cap}}(P)=
\frac{1}{C}\sum_{i=1}^{C}\mathrm{ReLU}\!\left(U^{\text{cpu}}_i(P)-X_i\right)
+ \\
\frac{1}{C}\sum_{i=1}^{C}\mathrm{ReLU}\!\left(U^{\text{ram}}_i(P)-D_i\right)
\end{multline}

\medskip
\noindent\textbf{Row-sum regularizer.}
We include a soft regularizer that encourages each CNF row to sum to one:
\begin{equation}
\mathcal{L}_{\text{rowsum}}(P)=\frac{1}{M}\sum_{m=1}^{M}\left(\sum_{i=1}^{C}P_{m,i}-1\right)^2.
\end{equation}
Note that, because $P$ is obtained via a row-wise softmax, this term is typically close to zero and acts mainly as a numerical sanity check.

\medskip
\noindent\textbf{Adjacency constraint.}
Let $c\in\mathbb{R}_+^{C\times C}$ be the cloud-to-cloud bandwidth matrix, and define the adjacency indicator $A_{i,j}=\mathbb{I}[c_{i,j}>0]$.
For each SFC hop edge $(u\!\to\!v)$ in the CNF chain graph, define the outer-product flow tensor
\begin{equation}
F^{(u,v)}_{i,j}(P)=P_{u,i}\,P_{v,j}.
\end{equation}
The adjacency penalty is the average probability mass assigned to non-existent links:
\begin{equation}
\mathcal{L}_{\text{adj}}(P)=
\frac{1}{|\mathcal{E}_{\text{sfc}}|\,C^2}
\sum_{(u,v)\in\mathcal{E}_{\text{sfc}}}\sum_{i=1}^{C}\sum_{j=1}^{C}
F^{(u,v)}_{i,j}(P)\,(1-A_{i,j}).
\end{equation}

\medskip
\noindent\textbf{Bandwidth constraint.}
Let $r_{u,v}$ be the communication rate (bps) required by hop $(u\!\to\!v)$.
The expected aggregate demand on physical link $(i,j)$ is
\begin{equation}
L_{i,j}(P)=\sum_{(u,v)\in\mathcal{E}_{\text{sfc}}} r_{u,v}\,F^{(u,v)}_{i,j}(P).
\end{equation}
We penalize any overuse with
\begin{equation}
\mathcal{L}_{\text{bw}}(P)=\frac{1}{C^2}\sum_{i=1}^{C}\sum_{j=1}^{C}\mathrm{ReLU}\!\left(L_{i,j}(P)-c_{i,j}\right).
\end{equation}

\begin{algorithm}[t]
\small
\caption{Training}
\label{alg:train}
\begin{algorithmic}[1]
\REQUIRE Dataset $\mathcal{D}$; heterogeneous graph builder $H(\cdot)$; mask $M$; steps $T$; schedule $\{\alpha_t\}_{t=1}^T$
\FOR{instances}
  \STATE Build graph $H$ and assignment $Y_0$
  \STATE Sample $t\!\sim\!\mathcal{U}\{1..T\}$, $\epsilon\!\sim\!\mathcal{N}(0,I)$; set $Y_t\!=\!\sqrt{\bar\alpha_t}Y_0+\sqrt{1-\bar\alpha_t}\,\epsilon$
  \STATE Predict $\hat\epsilon=\epsilon_\theta(H,\,Y_t,\,t)$
  \STATE Reconstruct $\widehat{Y}_0=(Y_t-\sqrt{1-\bar\alpha_t}\,\hat\epsilon)/\sqrt{\bar\alpha_t}$
  \STATE Apply placement mask: $\widehat{Y}_0\leftarrow \widehat{Y}_0\odot M + (-\infty)\cdot(1{-}M)$
  \STATE Row-wise probabilities $P=\mathrm{softmax}(\widehat{Y}_0)$
  \STATE Compute losses: $\mathcal{L}_{\text{DDPM}}=\|\epsilon-\hat\epsilon\|_2^2$; plus $\sum_u\lambda_u\mathcal{L}_u(P;H)$
  \STATE Update $\theta$ with Adam on $\mathcal{L}$
\ENDFOR
\end{algorithmic}
\end{algorithm}

\medskip
\noindent\textbf{End-to-end delay constraint}
For each hop $(u\!\to\!v)$, let $\tau_v$ be the processing delay of CNF $v$. We use a differentiable surrogate that distinguishes co-located and cross-node execution. Let
$\mathbb{I}[i=j]$ denote co-location and define a fixed message size $\kappa$ (in bits) used to model cross-node propagation as $\kappa/c_{i,j}$.
The expected per-hop delay is
\begin{multline}
\mathbb{E}[d_{u,v}](P)= 
\sum_{i=1}^{C}\sum_{j=1}^{C}F^{(u,v)}_{i,j}(P)\Big(
\mathbb{I}[i=j]\cdot\tau_v+ \\
\mathbb{I}[i\neq j]\cdot(\tau_v+\kappa/c_{i,j})
\Big).
\end{multline}
We sum these expected hop delays per service chain $h$ and add the processing delay of the last CNF once more, obtaining $D_h(P)$.
The penalty is then
\begin{equation}
\mathcal{L}_{\text{delay}}(P)=\frac{1}{H}\sum_{h=1}^{H}\mathrm{ReLU}\!\left(D_h(P)-\tau_h\right),
\end{equation}
where $\tau_h$ is the delay budget of chain $h$.

\medskip
Finally, the total training objective is
\begin{multline}
\mathcal{L}=
\mathcal{L}_{\text{denoise}}
+\lambda_{\text{cap}}\mathcal{L}_{\text{cap}}
+\lambda_{\text{restr}}\mathcal{L}_{\text{restr}}
+\lambda_{\text{rowsum}}\mathcal{L}_{\text{rowsum}}
+ \\
\lambda_{\text{adj}}\mathcal{L}_{\text{adj}}
+\lambda_{\text{bw}}\mathcal{L}_{\text{bw}}
+\lambda_{\text{delay}}\mathcal{L}_{\text{delay}}.
\end{multline}

Because all constraints act on $P$, our score network remains an unbiased estimator of the Gaussian noise while feasibility is enforced entirely in data space. The training and constraint losses procedure is summarized in Algorithm~\ref{alg:train}.

\subsection{Inference process}
\label{sec:Sampling}

After training, we use the learned GNN backbone to sample placements for new instances. 
The sampling procedure is the standard reverse diffusion (DDPM) schedule. We start 
from pure noise $Y_T\sim \mathcal{N}(0,I)$ of shape $1\times M\times C$ scaled by
$\sigma_T$, and, for $t=T-1,\dots,0$,  we iteratively predict and remove noise:
\begin{enumerate}
\item Given current $Y_t$ and noise level $\sigma_t$, compute $\hat{E} = 
\text{GNN}(H, Y_t, \sigma_t)$.

\item Estimate the underlying assignment 
$\hat{Y}_0 = (Y_t - \sigma_t\hat{E})/\sqrt{\bar\alpha_t}$.

\item If $t > 0$, compute the mean of $Y_{t-1}$ as $\mu = 
\sqrt{\bar\alpha_{t-1}} \, \hat{Y}_0$. After this compute, sample $Y_{t-1} = \mu + 
\sigma_{t-1} \zeta$, where $\zeta$ is fresh Gaussian noise.
\end{enumerate}

After reaching $t=0$, we have a denoised matrix $\hat{Y}_0$. We then apply the 
placement mask: forbidden entries of $\hat{Y}_0$ are set to $-\infty$, and finally 
we take a Softmax across clouds for each CNF to produce assignment probabilities.
This yields a matrix of probabilities $p_{m,i}$, where the highest-probability cloud 
is chosen for each CNF to form a discrete placement. In practice, we generate many 
samples and pick the one with lowest cost that satisfies all constraints. The best-of-$k$ sampling and selection used at inference is summarised in Algorithm~\ref{alg:infer}.

\begin{algorithm}[t]
\small
\caption{Inference}
\label{alg:infer}
\begin{algorithmic}[1]
\REQUIRE Instance graph $H$; mask $M$; steps $T$; schedule $\{\alpha_t\}$; samples $k$
\FOR{$s=1$ to $k$}
  \STATE Initialize $Y_T\sim\mathcal{N}(0,I)$
  \FOR{$t=T,\dots,1$}
     \STATE $\hat\epsilon=\epsilon_\theta(H,Y_t,t)$
     \STATE $\widehat{Y}_0=(Y_t-\sqrt{1-\bar\alpha_t}\,\hat\epsilon)/\sqrt{\bar\alpha_t}$; mask $\widehat{Y}_0\leftarrow \widehat{Y}_0\odot M + (-\infty)\cdot(1{-}M)$
     \STATE Compute mean $\mu_t$ of $p_\theta(Y_{t-1}\!\mid\!Y_t)$; sample $Y_{t-1}$
  \ENDFOR
  \STATE Discretize: $P=\mathrm{softmax}(\widehat{Y}_0)$; $\tilde{X}=\mathrm{argmax}_\text{row}(P)$
  \STATE Evaluate feasibility and cost of $\tilde{X}$
\ENDFOR
\STATE Return best feasible $\tilde{X}$ among $k$; if none feasible, return lowest-violation sample
\end{algorithmic}
\end{algorithm}

\subsection{Complexity of the solution}

Let $|V|$ and $|E|$ be the numbers of nodes and edges in the heterogeneous graph $H$, 
and let $L$ be the number of GNN layers. Each reverse step applies $L$ message-passing
layers whose cost is $O(|E|)$ (and $O(|V|)$ for node updates), hence one denoise step 
is $O(L\cdot(|E|{+}|V|))$. With $T$ diffusion steps and $k$ independent samples, the 
inference cost is $O\!\big(k\cdot T\cdot L\cdot (|E|{+}|V|)\big)$, which grows roughly
linearly with graph size and linearly in $T$. In contrast, MINLP solve time grows 
super-linearly and often exponentially with instance size due to branching and 
nonlinear constraints, leading to frequent timeouts on moderate/large cases. Our 
empirical results (Sec.~\ref{sec:results}) reflect this gap: diffusion runtime 
scales gently with $|C|$ while MINLP times increase steeply, with several instances 
hitting the time limit.

\section{Results}
\label{sec:results}

\subsection{Evaluation}\label{Evaluation}

The evaluation of the diffusion-based solver was conducted by generating $44$ network model scenarios, each characterised by distinct parameters, including the number of nodes and their respective resources, as well as the number of SFCs of varying lengths. The MINLP-based solver, created using GEKKO, is utilised in order to ascertain optimal solutions for the network models. The configuration is designed to identify the minimum-cost placement that is constrained-compliant. The system solves $39$ scenarios, of which 5 scenarios extend beyond the maximum time allowance of $4$ hours without yielding an optimal solution. The diffusion model, conversely, is utilised to generate $50$ solution samples for each instance, from which the optimal solution that fulfils all the constraints is selected.

For training and evaluating the diffusion-based solver, we split the $44$ network model
scenarios in two datasets:
\begin{itemize}
\item \textbf{Train dataset:} formed by $20$ network model scenarios, all with optimal 
solutions from the MINLP solver.
\item \textbf{Evaluation dataset:} formed by $24$ network model scenarios different from the 
train dataset, including the $5$ scenarios where the MINLP solver has not found any 
optimal solution.
\end{itemize}

With this set, we evaluate the diffusion-based solver and compare its performance to 
the MINLP-based solver in terms of solution feasibility, quality and efficiency. 
Our experiments are designed to answer the following questions: 
\begin{enumerate}
\item Does the diffusion model reliably generate feasible placements that meet 
all constraints?
\item How close are the diffusion model’s solutions to optimal in terms of total 
deployment cost?
\item What is the runtime advantage of the learned generative solver, especially as 
problem size scales?
\item Can the diffusion model generalize to scenarios that were not seen during 
training?
\end{enumerate}

In comparison with the previous version of this work, the evaluation in this paper is significantly extended in two directions. Firstly, the following additional baselines are introduced: two greedy heuristics (denoted Heuristic A and Heuristic B) and several learning-based solvers based on GDSG (Diffusion-GNN, Diffusion-MLP and a supervised GNN). Secondly, the definition of three additional evaluation families beyond the original validation set is proposed, with a view to assessing robustness and generalisation.

\begin{itemize}
    \item \emph{eval\_big}: larger instances with more nodes, CNFs and SFCs than those used 
    for training and validation.
    \item \emph{eval\_shift}: instances of similar size to the validation set but with 
    shifted placement costs and parameters, breaking the original Cloud/Fog/Edge 
    cost structure and stressing cost generalization.
    \item \emph{eval\_change}: a constraint-tight regime where effective message size and 
    link capacities are modified so that bandwidth and delay constraints become 
    strongly binding. This regime is designed to better reflect scenarios where 
    network resources are scarce and end-to-end SFC latency is critical.
\end{itemize}

In all cases, the same system model and feasibility checker is employed for all methods, thereby ensuring that every solver (MINLP, diffusion, heuristics and learning baselines) is evaluated under precisely the same constraints and cost function.

\subsection{Metrics}

We use the following key metrics to compare methods:
\begin{itemize}
\item \emph{Feasibility Rate:} The percentage of instances for which a method returns a solution that is both practical and achievable. It is an irrefutable fact that, by definition, whenever a solution is returned by the MINLP solver, it is always feasible. However, in the event that the solver fails to find a solution within the stipulated time limit, this is taken as evidence that the solver is not effectively feasible.

\item \emph{Total Deployment Cost:} The objective value of a placement that is feasible. In the validation set, the cost of each solver's placements is compared to the optimal cost from the MINLP. The present study reports both the average cost of the solutions and, when applicable, the relative optimality gap with respect to GEKKO, as defined as follows:
\[
\text{gap} = \frac{\text{cost} - \text{cost}_\text{GEKKO}}{\text{cost}_\text{GEKKO}},
\]
averaged over instances where both the method and GEKKO return feasible solutions.

\item \emph{Inference Time:} The temporal requirement for the production of a solution. In the context of the diffusion model, the process of inference entails the repeated execution of the GNN denoiser, $T=100$, on the specified instance, commencing from the initial state of Gaussian noise. The complexity of the inference process is largely independent of the complexity of the optimisation problem, with the exception of a linear relationship with the size of the input graph. In contradistinction, the MINLP solver's runtime exhibits a marked increase with problem size, approaching exponential growth in practice.

\item \emph{Regret and Best-share:} In the context of the extended evaluation families, where the GEKKO method is not utilised due to its computational expense, a comparative analysis is conducted between methodologies and the optimal cost among all approximate solvers for each individual instance. For the purpose of this study, let $c_i$ denote the cost achieved by method $m$ (if feasible), and let $c_i^m$ denote the cost achieved by the best method. We define the regret of method $m$ on that instance as
\[
r_i^m = \frac{c_i^m - c_i^\star}{c_i^\star},
\]
and report the average regret over instances where $m$ is feasible and $c_i^\star$ is 
defined. In addition, the \textit{best-share} of each method is reported. By this, it is meant the fraction of instances where the method attains the best cost (within a small numerical tolerance) among all methods.
\end{itemize}

The utilisation of these metrics enables the characterisation of two distinct aspects. Firstly, the feasibility of solutions that are discovered by a given method can be ascertained. Secondly, the proximity of these solutions to optimality (or to the best available baseline) can be determined. Furthermore, the computational cost of these solutions can be assessed.

\subsection{Results}
\begin{figure}[b]
\centering
\includegraphics[width=0.80\columnwidth]{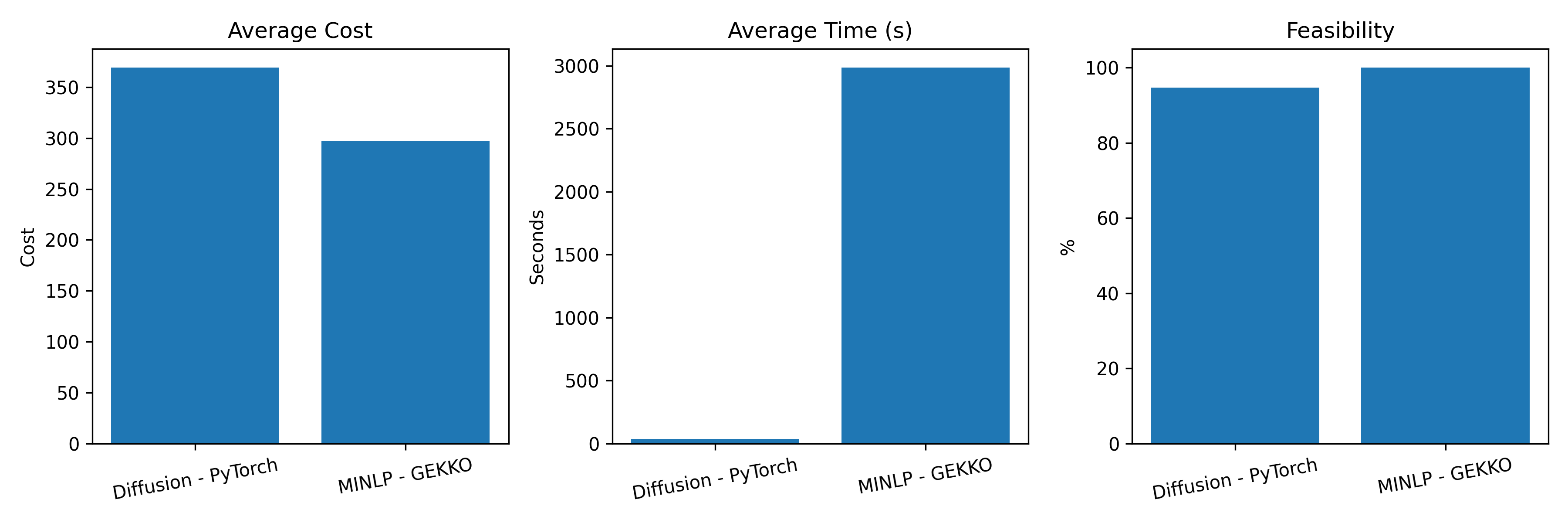}
\caption{Comparison between MINLP and diffusion solver.}
\label{fig:results}
\end{figure}

Figure~\ref{fig:results} summarizes the performance of the diffusion-based solver vs.\ 
the MINLP solver on representative test scenarios from the evaluation set.

In the first instance, the focus is on these results, and the analysis with our previous results~\cite{ourconferencepaper} is followed. The comparison is made of both solvers, with consideration given only to the evaluation dataset and the scenarios for which the MINLP solver finds an optimal solution. In~\cite{ourconferencepaper}, the scope of this figure was limited to a subset of 19 instances. In this expanded study, the focus has been broadened to encompass all 39 GEKKO-solvable instances. In summary, the diffusion-based solver successfully identifies feasible solutions in $92.3\%$ of the instances examined, thereby substantiating the efficacy of the model in identifying suboptimal solutions for network model scenarios, both those encountered during the training process and those that were not part of the training dataset. In terms of financial expenditure, the diffusion solver's best-found solution has an average cost of approximately $370$ units for solvable instances, in comparison to an average optimal cost of approximately $298$ units for the MINLP solver. This corresponds to an average optimality gap of approximately $25.6\%$ units. It is acknowledged that a key benefit of the diffusion model is its ability to generate a comprehensive distribution of solutions. This characteristic enables the trade-off between quality and runtime by producing multiple samples. In instances where the initial sample does not closely resemble the optimal solution, a modest number of samples often yields at least one solution that significantly surpasses the mean quality.

In terms of runtime, the results of the study demonstrate that the diffusion approach is significantly more efficient. On the GEKKO-solvable set, the MINLP solver requires an average of approximately $1957$ seconds per instance, whereas the diffusion solver produces $50$ samples in approximately $29.5$ seconds and can stop earlier if a good solution is found. As will be demonstrated in the following per-size analysis, this behaviour is also reflected therein.

In the instances where the MINLP solver encounters a timeout without yielding a solution, the diffusion-based solver identifies a suboptimal solution for all five scenarios. The metrics for these solutions are as follows: an average cost of $375$ units and an average time of $53$ seconds. This lends support to one of the key advantages of the diffusion-based solver over the MINLP solver: namely, its ability to identify suboptimal solutions for models with which the MINLP solver has been unable to find an optimal solution within the stipulated time limit.

\begin{figure}[b]
\centering
\includegraphics[width=0.90\columnwidth]{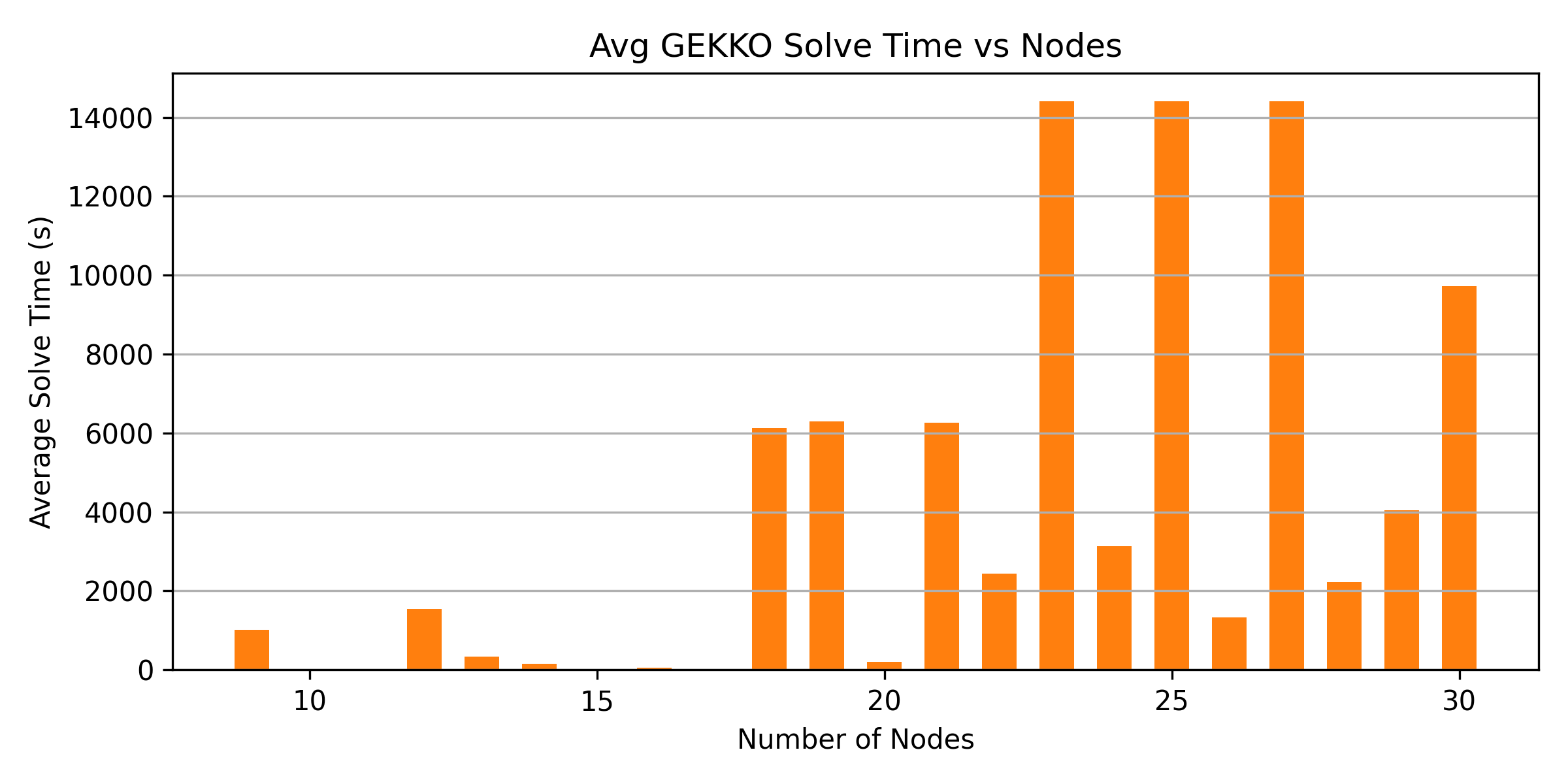}
\caption{Average time for MINLP solver vs.\ number of nodes.}
\label{fig:bar-gekko}
\end{figure}

\begin{figure}[b]
\centering
\includegraphics[width=0.90\columnwidth]{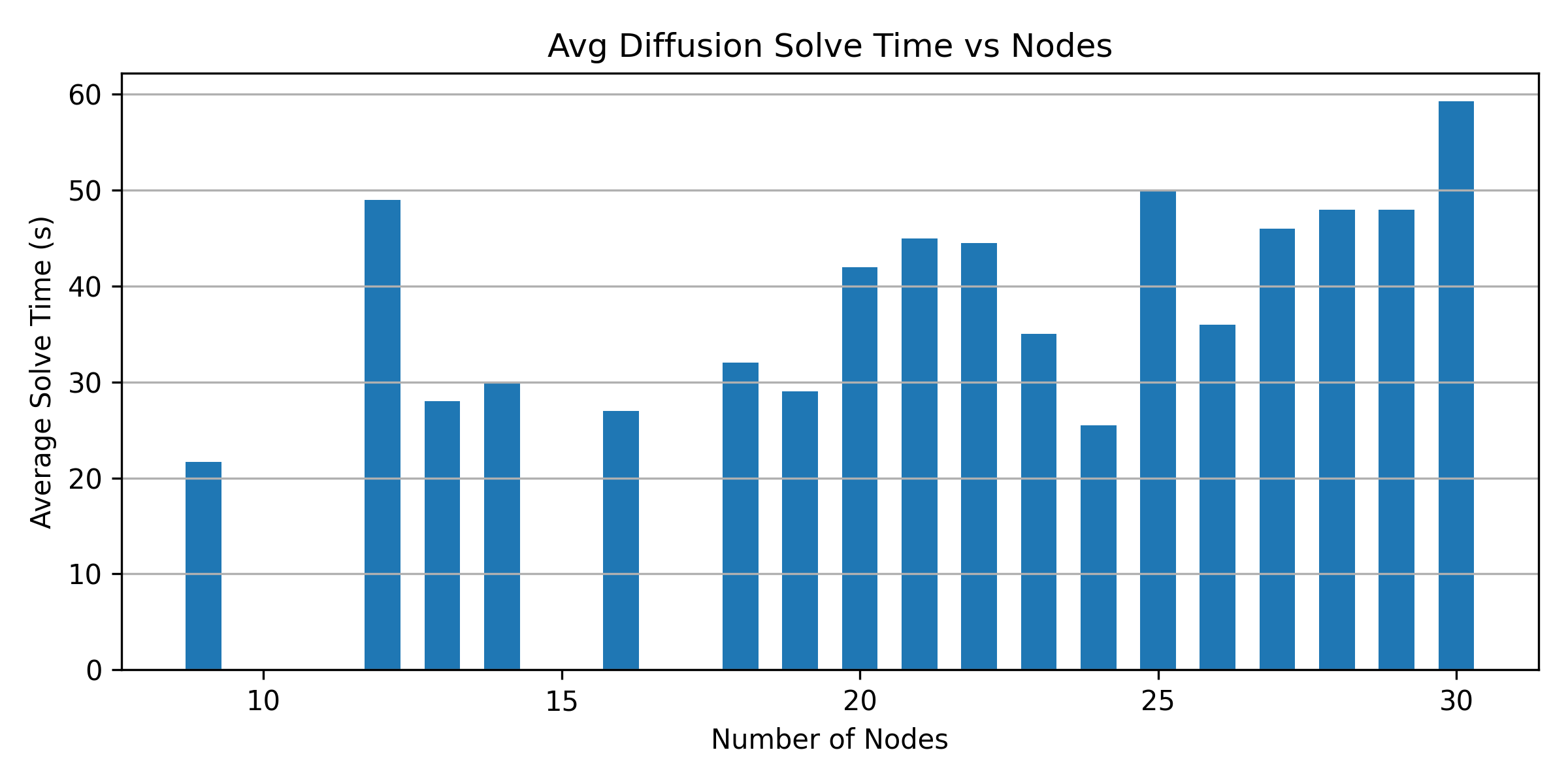}
\caption{Average time for diffusion-based solver vs.\ number of nodes.}
\label{fig:bar-diffusion}
\end{figure}

Finally, the time taken to find a solution for each solver was compared as a function of the number of nodes in the network model scenarios. The time was then averaged over scenarios with the same number of nodes. As demonstrated in Fig.~\ref{fig:bar-gekko}, the mean execution time for the MINLP solver exhibits a marked increase with the number of nodes, reaching a maximum of 14,400 seconds in instances that exceed the designated time limit. Conversely, as illustrated in Fig.~\ref{fig:bar-diffusion}, it is evident that the mean duration for the diffusion-based solver remains relatively constant as the number of nodes increases. This observation validates the hypothesis that the solver's execution time remains largely unaffected by the combinatorial intricacy of the optimization problem, exhibiting a approximate linear relationship with the graph size.

\subsection{Extended baseline comparison}

\begin{table*}[t]
\centering
\caption{Overall comparison on the base benchmark (44 scenarios)}
\label{tab:base-benchmark}
\footnotesize
\setlength{\tabcolsep}{3.5pt}
\begin{tabular}{lccccccc}
\toprule
& \multicolumn{4}{c}{GEKKO-solvable set (N=39)} & \multicolumn{3}{c}{GEKKO-timeout set (N=5)}\\
\cmidrule(lr){2-5}\cmidrule(lr){6-8}
Method & Feas. (\%) & Cost & Gap (\%) & Time (s) & Feas. (\%) & Cost & Time (s)\\
\midrule
MINLP--GEKKO                    & 100.0 & 298.46 &  0.00 & 1956.756 &   0.0 & --     & 14400.000\\
(own) Diffusion-PyTorch              &  92.3 & 369.86 & 25.60 &   29.487 & 100.0 & 375.00 &    53.000\\
Diffusion-GNN (GDSG)            &   2.6 & 210.00 &  0.00 &   37.238 &   0.0 & --     &    45.774\\
Diffusion-MLP (GDSG)            &   7.7 & 220.00 &  0.00 &    3.253 &   0.0 & --     &     3.103\\
Supervised GNN (GDSG)           &   2.6 & 210.00 &  0.00 &    0.009 &   0.0 & --     &     0.012\\
Heuristic A (SFC-centric)       &  71.8 & 329.29 & 13.10 &    0.024 &  60.0 & 361.67 &     0.090\\
Heuristic B (node+BW greedy)    &  92.3 & 298.06 &  0.20 &    0.011 & 100.0 & 319.00 &     0.024\\
\bottomrule
\end{tabular}
\end{table*}

The extended evaluation comprises supplementary baselines that extend beyond the MINLP and diffusion solvers. In this study, the focus is on two greedy heuristics (Heuristic A and Heuristic B) and three learning-based solvers derived from the GDSG framework. These solvers include an unsupervised Diffusion-GNN, an unsupervised Diffusion-MLP, and a supervised GNN. Table~\ref{tab:base-benchmark} summarizes the aggregate performance on the original 44-scenario benchmark, separating GEKKO-solvable instances from GEKKO timeouts.

The set under consideration is designated as the GEKKO-solvable set. In 39 out of the total number of instances, the MINLP solver achieves $100\%$ feasibility and $0\%$ gap by construction, with an average solve time of approximately $1956.8$ seconds per instance. The diffusion-based solver achieves a feasibility rate of $92.3\%$ on these instances, with an average cost of $369.9$ units and an average optimality gap of $25.6\%$.

The two heuristics demonstrate markedly divergent behavioural patterns. It is evident that Heuristic A, which is SFC-centric, identifies feasible solutions for 71.8\% of the GEKKO-solvable instances. The average cost of these solutions is $329.3$ units, while the average gap is $13.1\%$. It is noteworthy that these processes exhibit negligible runtime, with an average of $24$ milliseconds per instance. In contrast, Heuristic B, which is node- and bandwidth-aware, exhibits a combination of high feasibility (92.3\% match with diffusion on this set) with an average cost of 298.1 units and an average gap of only 0.2\%. Furthermore, it operates with a processing time of approximately 11 milliseconds per instance. In summary, the Heuristic B is, in essence, optimal with respect to the GEKKO benchmark. Furthermore, it is more than three orders of magnitude faster than MINLP and approximately three orders of magnitude faster than diffusion.

It is evident that the learning baselines are non-competitive as solvers. The unsupervised Diffusion-GNN and the supervised GNN are only feasible on 2.6\% of the GEKKO-solvable instances (one instance out of 39), while the Diffusion-MLP reaches 7.7\% feasibility (three instances). In the limited number of instances where they are viable, these solutions correspond to the GEKKO optimum yielding an average discrepancy of $0\%$, yet they consistently fall short in producing valid solutions for the majority of cases. As a result, they cannot be regarded as practical solvers for this problem.

In the context of the five GEKKO time-limited instances, the observed behaviour aligns with the previous version~\cite{ourconferencepaper}. This indicates that both the heuristics and the diffusion solver possess the capability to swiftly generate feasible solutions, while GEKKO reaches the 4~h limit. It is evident that Heuristic B attains $100\text{th}$ feasibility with an average cost of $319$ units and $24$ ms runtime, whereas the diffusion solver attains $100\text{th}$ feasibility with an average cost of $375$ units and $53$s runtime.

\medskip
\noindent\textbf{Difficulty-bucket analysis.} 

In order to facilitate a more profound comprehension of the manner in which performance varies in accordance with problem difficulty, a stratification of the 39 GEKKO-solvable instances has been undertaken, resulting in the establishment of three difficulty buckets (easy, medium, hard) based on their respective GEKKO solve times, with 13 instances allocated to each bucket. The feasibility of each bucket, along with its average gap and average time, is computed. Table~\ref{tab:bucket-feas-gap-time} reports feasibility, optimality gap and average runtime across buckets, highlighting how constraint satisfaction and suboptimality evolve with instance difficulty.

\begin{table*}[t]
\centering
\caption{Feasibility, optimality gap, and runtime by difficulty bucket on the GEKKO-solvable set (N=39)}
\label{tab:bucket-feas-gap-time}
\footnotesize
\setlength{\tabcolsep}{5pt}
\begin{tabular}{lccccccccc}
\toprule
& \multicolumn{3}{c}{Easy (N=13)} & \multicolumn{3}{c}{Medium (N=13)} & \multicolumn{3}{c}{Hard (N=13)} \\
\cmidrule(lr){2-4}\cmidrule(lr){5-7}\cmidrule(lr){8-10}
Method
& Feas. (\%) & Gap (\%) & Time (s)
& Feas. (\%) & Gap (\%) & Time (s)
& Feas. (\%) & Gap (\%) & Time (s) \\
\midrule
MINLP--GEKKO                 & 100.0 &  0.00 &  585.900 & 100.0 &  0.00 & 1608.742 & 100.0 &  0.00 & 3675.627 \\
Diffusion--PyTorch           &  92.3 & 22.40 &   22.923 &  84.6 & 25.02 &   24.923 & 100.0 & 28.99 &   40.615 \\
Diffusion-GNN (GDSG)         &   0.0 &   --  &   29.653 &   7.7 &  0.00 &   30.021 &   0.0 &   --  &   52.040 \\
Diffusion-MLP (GDSG)         &   0.0 &   --  &    3.748 &   7.7 &  0.00 &    3.014 &  15.4 &  0.00 &    2.997 \\
Supervised GNN (GDSG)        &   0.0 &   --  &    0.008 &   7.7 &  0.00 &    0.010 &   0.0 &   --  &    0.010 \\
Heuristic A (SFC-centric)    &  76.9 & 15.13 &    0.011 &  61.5 & 14.57 &    0.014 &  76.9 &  9.82 &    0.048 \\
Heuristic B (node+BW greedy) &  92.3 &  0.22 &    0.006 &  84.6 &  0.00 &    0.006 & 100.0 &  0.35 &    0.021 \\
\bottomrule
\end{tabular}
\end{table*}

In the instances classified as \textit{easy}, Heuristic B has been shown to demonstrate near-optimal performance. It achieves $92.3\%$ feasibility with an average gap of $0.22\%$ and a runtime of $6$ ms. Meanwhile, the diffusion solver attains $92.3\%$ feasibility but experiences a $22.4\%$ gap and requires approximately $22.9$s per instance. As demonstrated in Figure 1, the gap in diffusion increases to 25.0\% for medium instances, with a feasibility rate of 84.6\%. In contrast, Heuristic B maintains a feasibility rate of 84.6\% and a gap of 0\%, aligning with the performance of GEKKO on all medium instances that are feasible. 

In the cases of greater complexity, diffusion still attains 100\% feasibility, but with a 29.0\% gap and an approximate runtime of 40.6 seconds. Conversely, Heuristic B attains 100\% feasibility and a mere 0.35\% gap in approximately 21 milliseconds. In terms of all the criteria under consideration, Heuristic B is shown to be the optimal solution, with a high level of efficiency. In contrast, the diffusion solver is found to be substantially suboptimal in terms of cost, and is significantly slower than the heuristics.

The learning baselines continue to demonstrate an almost complete lack of feasibility across all buckets (0–15\% feasibility), thereby reinforcing the conclusion that they do not constitute effective solutions for this problem.

The ensuing results generalise the findings previously outlined in the previous paper~\cite{ourconferencepaper}. With regard to the original instance distribution, a carefully engineered greedy heuristic (Heuristic B) is shown to be highly effective, essentially matching the MINLP optimum even on the most challenging instances. In contrast, the diffusion-based solver is found to be significantly suboptimal in terms of cost and considerably slower than the heuristics.

\medskip
\noindent\textbf{Robustness to scale and parameter shifts: eval\_big and eval\_shift}

We next evaluate robustness under two types of distribution shift beyond the validation set considered in the previous version:

\begin{itemize}
    \item \textbf{eval\_big:} larger instances with more CNFs and SFCs than those 
    used for training and validation.
    \item \textbf{eval\_shift:} instances with the same size distribution as the 
    validation set but with perturbed placement costs and parameters.
\end{itemize}

It is not possible to run GEKKO for these families due to the computational cost of solving many large instances. Consequently, the evaluation of methods is undertaken by means of relative regret and best-share with respect to the best cost achieved by any approximate solver on each instance. It is imperative to note that the statistics exclusively encompass instances wherein at least one solver successfully identifies a feasible solution. Table~\ref{tab:ood-big-shift-change} reports feasibility, regret, and best-share on the three out-of-distribution families.

\begin{table*}[t]
\centering
\caption{Relative-cost statistics on scale/parameter-shift and constraint-tight families.}
\label{tab:ood-big-shift-change}
\footnotesize
\setlength{\tabcolsep}{4pt}
\begin{tabular}{llcccc}
\toprule
Family & Method & Feas. (\%) & Avg. regret (\%) & Best-share (\%) & Time (s) \\
\midrule
eval\_big (N=59)   & Diffusion--PyTorch              &  25.4 & 31.61 &   0.0 & 58.068 \\
eval\_big (N=59)   & Heuristic A (SFC-centric)       &  86.4 &  6.29 &  15.3 &  0.323 \\
eval\_big (N=59)   & Heuristic B (node+BW greedy)    & 100.0 &  0.00 & 100.0 &  0.030 \\
\midrule
eval\_shift (N=55) & Diffusion--PyTorch              &  98.2 & 47.34 &   3.6 & 27.582 \\
eval\_shift (N=55) & Heuristic A (SFC-centric)       &  80.0 & 27.24 &  14.5 &  0.107 \\
eval\_shift (N=55) & Heuristic B (node+BW greedy)    &  98.2 &  0.00 &  98.2 &  0.029 \\
\midrule
eval\_change (N=35) & Diffusion--PyTorch              & 91.4 &  8.65 & 57.1 & 35.457 \\
eval\_change (N=35) & Heuristic A (SFC-centric)       & 17.1 &  0.00 & 17.1 &  0.041 \\
eval\_change (N=35) & Heuristic B (node+BW greedy)    & 31.4 &  0.18 & 28.6 &  0.014 \\
\bottomrule
\end{tabular}
\end{table*}

\begin{figure}[b]
\centering
\includegraphics[width=0.95\columnwidth]{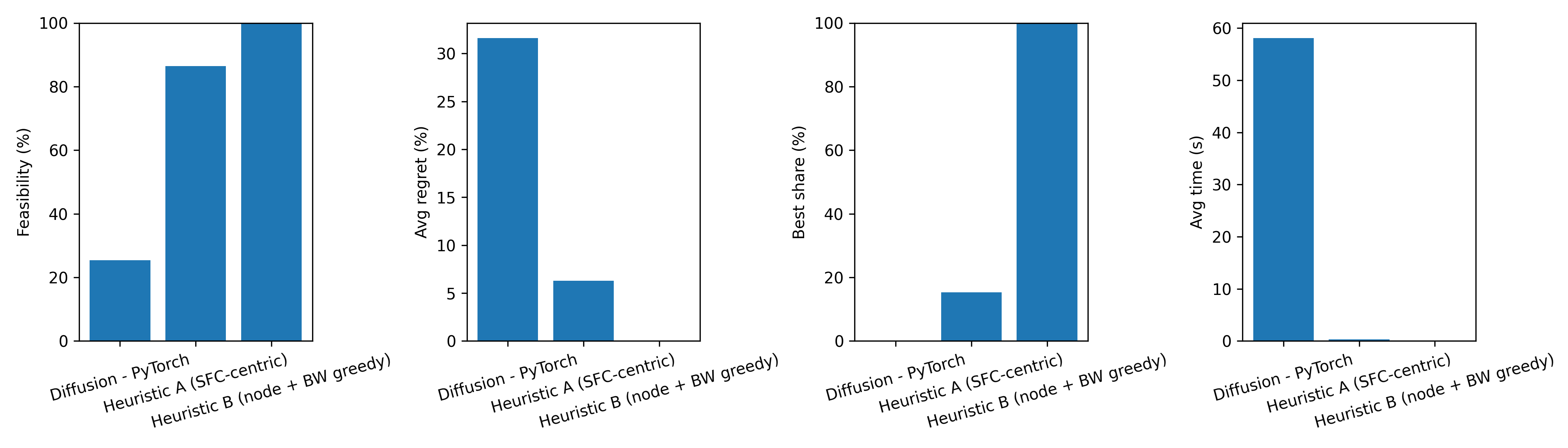}
\caption{Relative-cost evaluation on \texttt{eval\_big}: feasibility, regret, best-share and runtime.}
\label{fig:eval-big}
\end{figure}

On \textbf{eval\_big} ($N=59$ instances with at least one feasible solution), 
Heuristic B clearly dominates, as can be seen in Fig.~\ref{fig:eval-big}:
\begin{itemize}
    \item Heuristic B achieves $100\%$ feasibility, $0\%$ average regret and 
    best-share $=100\%$, with an average runtime of $30$~ms.
    \item Heuristic A attains $86.4\%$ feasibility, $6.29\%$ average regret and 
    best-share $=15.3\%$, with $0.323$~s average runtime.
    \item Diffusion–PyTorch attains only $25.4\%$ feasibility, with $31.61\%$ 
    average regret and best-share $=0\%$, requiring about $58.1$~s per instance.
\end{itemize}

It is evident that, on larger graphs, the greedy heuristic B exhibits both robustness and near-optimality. Conversely, the diffusion solver's feasibility is subject to collapse, and in instances of feasibility, its cost is significantly higher than that of the optimal solver.

\begin{figure}[b]
\centering
\includegraphics[width=0.95\columnwidth]{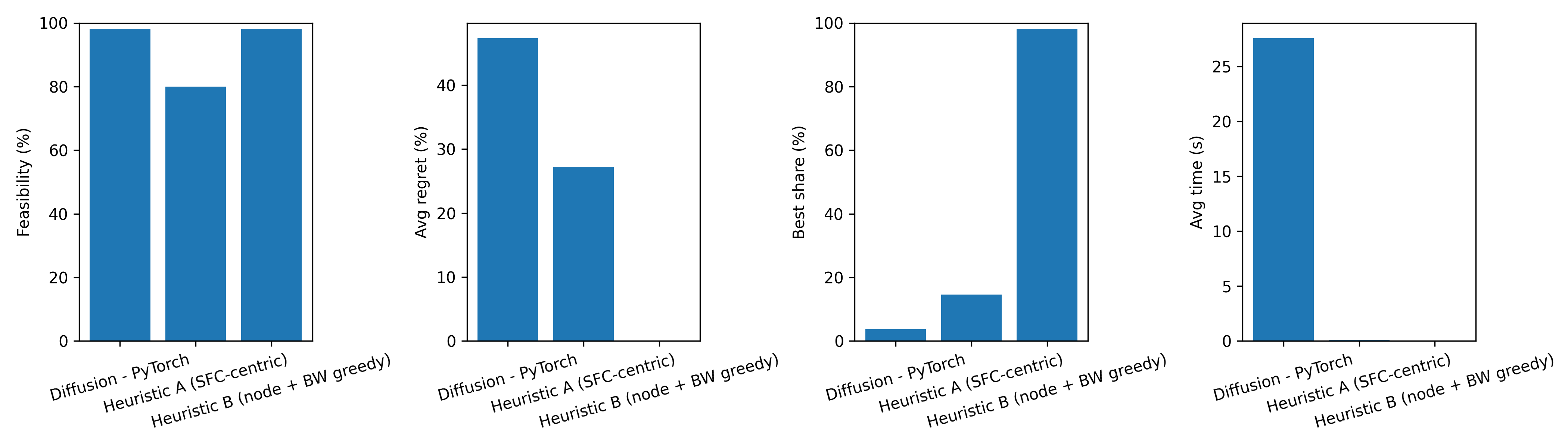}
\caption{Relative-cost evaluation on \texttt{eval\_shift}: feasibility, regret, best-share and runtime.}
\label{fig:eval-shift}
\end{figure}

On \textbf{eval\_shift} ($N=55$), where we modify cost and parameter distributions 
while keeping problem size similar to the validation set, the picture is similar, as can be seen in Fig.~\ref{fig:eval-shift}:
\begin{itemize}
    \item Heuristic B attains $98.2\%$ feasibility, $0\%$ average regret and 
    best-share $=98.2\%$, with about $29$~ms runtime.
    \item Heuristic A reaches $80.0\%$ feasibility, $27.24\%$ average regret and 
    best-share $=14.5\%$, with $0.107$~s runtime.
    \item Diffusion–PyTorch maintains $98.2\%$ feasibility, but suffers an average 
    regret of $47.34\%$ and best-share $=3.6\%$, with $27.6$~s runtime.
\end{itemize}

The experiments conducted confirm the hypothesis that strong domain-specific heuristics can be extremely difficult to surpass. Heuristic B demonstrates excellent performance under both size and cost/parameter shifts, while the diffusion solver exhibits a significant degradation in performance.

\medskip
\noindent\textbf{Constraint-tight regime: eval\_change}

The most intriguing behaviour manifests in the eval\_change family, wherein we deliberately constrain bandwidth and delay parameters to more accurately emulate a scenario characterised by limited network resources and the significance of end-to-end SFC latency. In practice, the effective message size employed for the calculation of per-hop network delay and link load is increased, whilst link capacities are reduced. This ensures that both bandwidth and delay constraints become strongly binding. It is precisely in this regime that local, greedy decisions are expected to become less reliable, and that global patterns learned from GEKKO solutions may prove advantageous. Table~\ref{tab:ood-big-shift-change} quantifies the regime shift where feasibility becomes the dominant differentiator.

\begin{figure}[b]
\centering
\includegraphics[width=0.95\columnwidth]{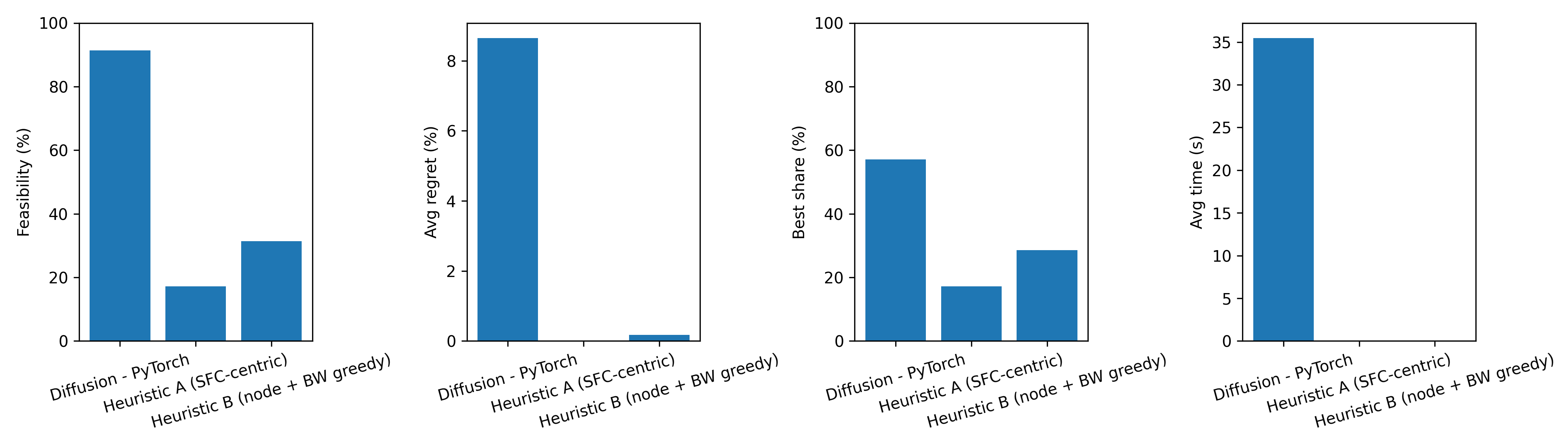}
\caption{Relative-cost evaluation on \texttt{eval\_change}: feasibility, regret, best-share and runtime.}
\label{fig:eval-change}
\end{figure}

On \textbf{eval\_change} ($N=35$ instances with at least one feasible solution), the 
results are as follows, as can be seen in Fig.~\ref{fig:eval-change}:
\begin{itemize}
    \item Diffusion–PyTorch achieves $91.4\%$ feasibility, with an average regret of 
    $8.65\%$, best-share $=57.1\%$ and average runtime of $35.5$~s.
    \item Heuristic B achieves $31.4\%$ feasibility, with an average regret of 
    $0.18\%$, best-share $=28.6\%$ and runtime $14$~ms.
    \item Heuristic A achieves $17.1\%$ feasibility, with $0\%$ average regret and 
    best-share $=17.1\%$, and average runtime $41$~ms.
    \item All three learning-based solvers (Diffusion-GNN, Diffusion-MLP and 
    supervised GNN) obtain $0\%$ feasibility on these instances, despite consuming 
    non-negligible runtime.
\end{itemize}

The following observations are of particular significance. Initially, Heuristic~B, which demonstrated optimal performance in the original validation set, exhibits fragility within this constraint-tight regime. It successfully generates valid solutions in approximately one-third of the instances. In instances where this is a viable option, the cost remains predominantly optimal (average regret $0.18\%$), and frequently attains the most optimal cost (best-share $28.6\%$). However, in approximately $70\%$ of cases, it is unsuccessful in identifying a valid placement. It is evident that Heuristic~A is even more fragile, with a feasibility percentage of only 17.1\%.

Secondly, the diffusion-based solver demonstrates significantly greater robustness. It maintains high feasibility ($91.4\%$) in this challenging regime and attains optimal cost on 57.1\% of the instances. The mean regret of approximately 8.7\% of the optimal solution per instance is moderate, particularly given that the \textit{optimal solution} is occasionally the heuristic and at other times diffusion itself.

Thirdly, the learning baselines demonstrate a complete collapse, with no capacity to generate a single viable solution on eval\_change, despite having been trained on the original validation regime. Evidence suggests that subjects demonstrate substandard performance not only in relation to the base distribution, but also in terms of their ability to adapt to the tightened constraints.

\begin{figure}[b]
\centering
\includegraphics[width=0.85\columnwidth]{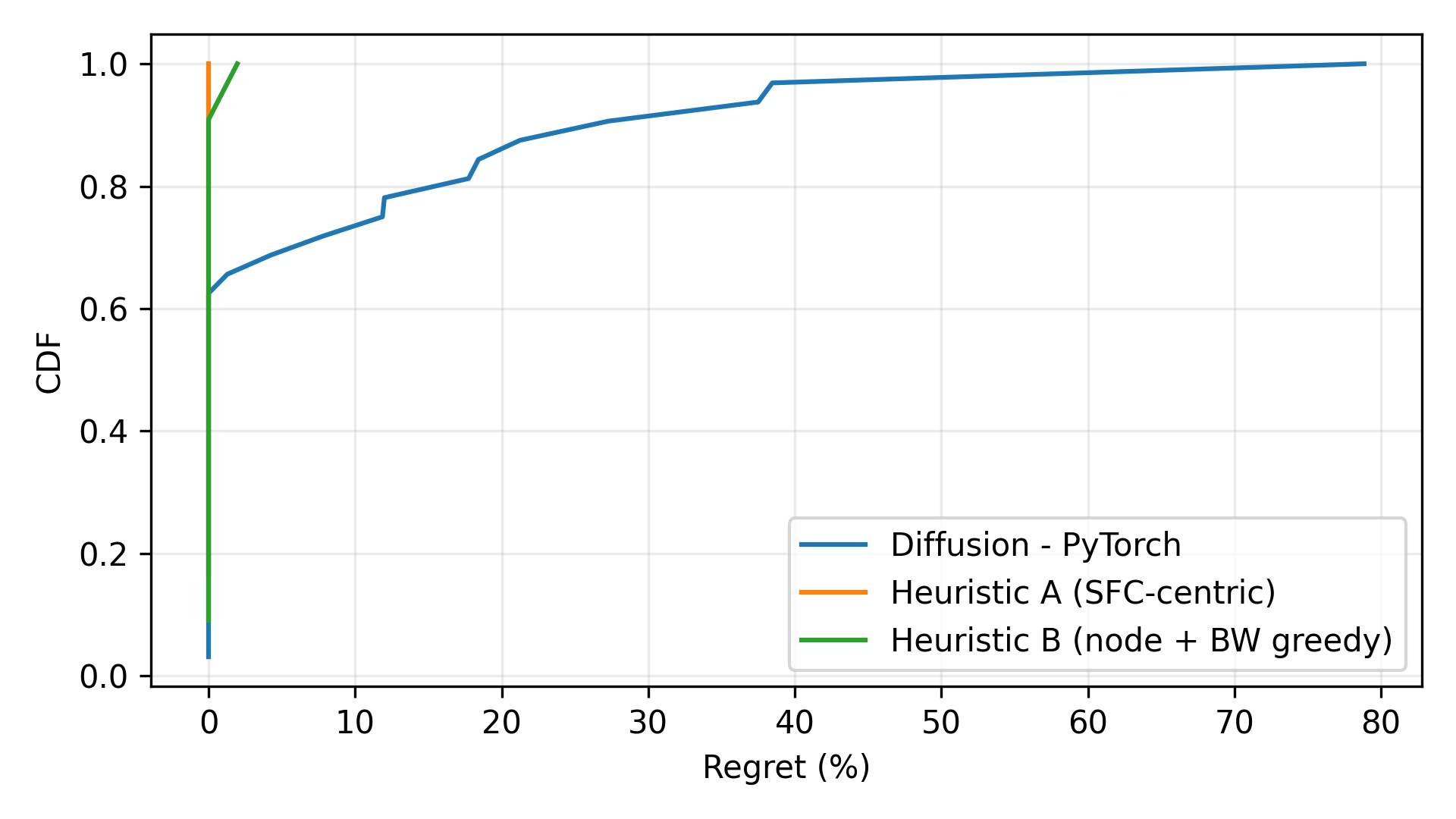}
\caption{CDF of regret on \texttt{eval\_change}.}
\label{fig:cdf-eval-change}
\end{figure}

To complement average regret values with a distributional view, Fig.~\ref{fig:cdf-eval-change} reports the empirical CDF of regret on \texttt{eval\_change} for all feasible solutions returned by each method. This figure highlights not only which solver attains lower regret on average, but also how consistently each method stays close to the best solution across instances.

In summary, the evaluation on the eval\_change dataset reveals a regime where the proposed diffusion-based solver outperforms the best heuristic when both feasibility and cost are considered. The heuristic remains near-optimal whenever it succeeds, but its success probability is low. In contrast, diffusion is feasible on almost all instances and often yields the best solution among all methods.

\subsection{Discussion}


The amalgamation of all experiments has resulted in the following key conclusions being supported by the extended evaluation. As demonstrated in~\cite{ourconferencepaper}, the original distribution, as well as larger instances (eval\_big) and moderate cost shifts (eval\_shift), employ a carefully engineered greedy heuristic (Heuristic~B) that is of particular note. This heuristic attains near-optimal cost with very high feasibility and negligible runtime. In the aforementioned regimes, it has been demonstrated that the diffusion solver is significantly suboptimal in terms of cost. This finding calls into question the justification for its additional complexity and runtime.

It is evident that simple learning baselines (GDSG with GNN/MLP and supervised GNN) are not viable as solvers. These baselines produce feasible solutions on only a small percentage of instances in the base distribution, and they are unable to function effectively under the more challenging regime (eval\_change).

However, when transitioning to a constraint-tight regime, characterised by stringent bandwidth and delay constraints (eval\_change), the scenario undergoes a transformation. The heuristic, which previously exhibited optimal performance, becomes susceptible to disruption. Conversely, the proposed diffusion-based solver demonstrates a heightened degree of robustness, ensuring feasibility in approximately 91\% of instances and attaining optimal cost in about 57\% of them, accompanied by a moderate average regret.

The results obtained provide a compelling illustration of the underlying dynamics. In instances where a robust heuristic is in place and the constraints are relatively flexible, diffusion-based solvers encounter significant challenges in their ability to compete. However, when constraints are tightened and the global structure of the solution becomes more important than local, greedy decisions, a diffusion-based solver trained on optimal placements can provide robust, high-quality solutions where handcrafted rules break down. This is precisely the type of setting in which generative solvers, such as diffusion models, are considered to be advantageous for network optimisation. These solvers are not intended to replace simple heuristics in straightforward scenarios; rather, they are regarded as robust solutions in complex, constrained contexts where traditional heuristics are ineffective.

\section{Conclusion}
\label{sec:conclusions}

In summary, we have adapted the DDPM framework to the CNF/SFC placement problem by:
\begin{enumerate}
    \item encoding the network and CNFs as a graph input,
    \item designing a GNN that predicts the Gaussian noise required for denoising steps,
    \item integrating domain constraints as losses.
\end{enumerate}

Each component is grounded in diffusion theory and network optimization 
literature. The use of a GraphSAGE encoder with edge features ensures node capacities 
and link data influence the embedding. The time embedding follows standard DDPM practice,
allowing the network to calibrate its denoising at each step. The MLP decoder on 
concatenated pair embeddings is analogous to score networks in continuous DDPMs, but 
adapted here to output a score for each discrete choice.

This diffusion-based approach leverages the strengths of neural generative modelling 
for combinatorial optimization. The iterative denoising process allows the model to 
gradually refine a placement and effectively navigating the exponentially large 
solution space. By using a GNN, it captures the rich structure of the network and CNFs 
embedding global state in the latent vectors. Unlike greedy heuristics or standalone 
neural predictors, the diffusion model can produce diverse solutions by sampling, which 
is useful since the problem is NP-hard and multiple good solutions may exist.

In conclusion, the diffusion-based CNF placement solver demonstrates how generative
modelling and graph neural networks can be combined to tackle network optimization. 
It bridges the theoretical foundations of DDPMs with the specific constraints of CNFs, 
yielding a practical end-to-end system.

\subsection{Limitations}
It is important to note that this study is also subject to limitations. Firstly, the quality of the solution is contingent upon best-of-$k$ sampling, a process which engenders a controllable yet non-negligible inference cost as $k$ increases. Secondly, the present training system is overseen by optimal (or near-optimal) solver-generated placements, which can be costly to acquire on larger scales. Incorporating training from suboptimal data or self-improvement loops is a logical subsequent step. Thirdly, the focus of this evaluation is on a static, cloud-only continuum and sequential-chain delay accounting. However, extending this to encompass richer continuum dynamics (e.g., time-varying links, multi-hop forwarding, and full DAG delay semantics) would serve to broaden its applicability.

\subsection{Future work}
Future research will examine the utilisation of more robust constraint satisfaction mechanisms during the sampling process, such as guidance, projection, or repair. Additionally, it will explore architectural variants that enhance cost optimality in scenarios where strong greedy heuristics already prevail.

\bibliographystyle{IEEEtran}
\bibliography{DiffusionCNF}

@Article{liang2025diffsg,
  author    = {Liang, Ruihuai and Yang, Bo and Yu, Zhiwen and Guo, Bin and Cao, Xuelin and Debbah, Mérouane and Poor, H. Vincent and Yuen, Chau},
  journal   = {IEEE Communications Magazine},
  title     = {DiffSG: A Generative Solver for Network Optimization with Diffusion Model},
  year      = {2025},
  issn      = {1558-1896},
  month     = jun,
  number    = {6},
  pages     = {16--24},
  volume    = {63},
  doi       = {10.1109/mcom.001.2400428},
  publisher = {Institute of Electrical and Electronics Engineers (IEEE)},
}

@InProceedings{zhang,
  author    = {Zhang, Zuyuan and Aggarwal, Vaneet and Lan, Tian},
  booktitle = {IEEE INFOCOM 2025 - IEEE Conference on Computer Communications},
  title     = {Network Diffuser for Placing-Scheduling Service Function Chains with Inverse Demonstration},
  year      = {2025},
  month     = may,
  pages     = {1--10},
  publisher = {IEEE},
  doi       = {10.1109/infocom55648.2025.11044702},
}

@Article{satpathy,
  author    = {Satpathy, Anurag and Sahoo, Manmath Narayan and Swain, Chittaranjan and Bellavista, Paolo and Guizani, Mohsen and Muhammad, Khan and Bakshi, Sambit},
  journal   = {IEEE Communications Surveys \& Tutorials},
  title     = {Virtual Network Embedding: Literature Assessment, Recent Advancements, Opportunities, and Challenges},
  year      = {2025},
  issn      = {2373-745X},
  pages     = {1--1},
  doi       = {10.1109/comst.2025.3531724},
  publisher = {Institute of Electrical and Electronics Engineers (IEEE)},
}

@Article{liangexplorations,
  author    = {Liang, Ruihuai and Yang, Bo and Chen, Pengyu and Li, Xianjin and Xue, Yifan and Yu, Zhiwen and Cao, Xuelin and Zhang, Yan and Debbah, Mérouane and Vincent Poor, H. and Yuen, Chau},
  journal   = {IEEE Internet of Things Journal},
  title     = {Diffusion Models as Network Optimizers: Explorations and Analysis},
  year      = {2025},
  issn      = {2372-2541},
  month     = may,
  number    = {10},
  pages     = {13183--13193},
  volume    = {12},
  doi       = {10.1109/jiot.2025.3528955},
  publisher = {Institute of Electrical and Electronics Engineers (IEEE)},
}

@Article{hdu,
  author    = {Du, Hongyang and Zhang, Ruichen and Liu, Yinqiu and Wang, Jiacheng and Lin, Yijing and Li, Zonghang and Niyato, Dusit and Kang, Jiawen and Xiong, Zehui and Cui, Shuguang and Ai, Bo and Zhou, Haibo and Kim, Dong In},
  journal   = {IEEE Communications Surveys \&Tutorials},
  title     = {Enhancing Deep Reinforcement Learning: A Tutorial on Generative Diffusion Models in Network Optimization},
  year      = {2024},
  issn      = {2373-745X},
  number    = {4},
  pages     = {2611--2646},
  volume    = {26},
  doi       = {10.1109/comst.2024.3400011},
  publisher = {Institute of Electrical and Electronics Engineers (IEEE)},
}

@Article{GDSG,
  author    = {Liang, Ruihuai and Yang, Bo and Chen, Pengyu and Cao, Xuelin and Yu, Zhiwen and Debbah, Mérouane and Niyato, Dusit and Poor, H. Vincent and Yuen, Chau},
  journal   = {IEEE Transactions on Mobile Computing},
  title     = {GDSG: Graph Diffusion-Based Solution Generator for Optimization Problems in MEC Networks},
  year      = {2025},
  issn      = {2161-9875},
  month     = oct,
  number    = {10},
  pages     = {10264--10277},
  volume    = {24},
  doi       = {10.1109/tmc.2025.3568248},
  publisher = {Institute of Electrical and Electronics Engineers (IEEE)},
}

@InProceedings{difusco,
  author    = {Sun, Zhiqing and Yang, Yiming},
  booktitle = {Proceedings of the 37th International Conference on Neural Information Processing Systems},
  title     = {DIFUSCO: graph-based diffusion solvers for combinatorial optimization},
  year      = {2023},
  address   = {Red Hook, NY, USA},
  publisher = {Curran Associates Inc.},
  series    = {NIPS '23},
  articleno = {164},
  location  = {New Orleans, LA, USA},
  numpages  = {26},
}

@Article{Doan,
  author    = {Doan, Khai and Avgeris, Marios and Leivadeas, Aris and Lambadaris, Ioannis and Shin, Wonjae},
  journal   = {IEEE Transactions on Vehicular Technology},
  title     = {Cooperative Learning-Based Framework for VNF Caching and Placement Optimization Over Low Earth Orbit Satellite Networks},
  year      = {2025},
  issn      = {1939-9359},
  month     = mar,
  number    = {3},
  pages     = {4758--4773},
  volume    = {74},
  doi       = {10.1109/tvt.2024.3487015},
  publisher = {Institute of Electrical and Electronics Engineers (IEEE)},
}

@InProceedings{ddpm,
  author    = {Ho, Jonathan and Jain, Ajay and Abbeel, Pieter},
  booktitle = {Proceedings of the 34th International Conference on Neural Information Processing Systems},
  title     = {Denoising diffusion probabilistic models},
  year      = {2020},
  address   = {Red Hook, NY, USA},
  publisher = {Curran Associates Inc.},
  series    = {NIPS '20},
  articleno = {574},
  isbn      = {9781713829546},
  location  = {Vancouver, BC, Canada},
  numpages  = {12},
}

@inproceedings{
  song2021scorebased,
  title={Score-Based Generative Modeling through Stochastic Differential Equations},
  author={Yang Song and Jascha Sohl-Dickstein and Diederik P Kingma and Abhishek Kumar and Stefano Ermon and Ben Poole},
  booktitle={International Conference on Learning Representations},
  year={2021},
  url={https://openreview.net/forum?id=PxTIG12RRHS}
}

@techreport{etsi_nfv_mano,
  author       = {{ETSI}},
  title        = {{Network Functions Virtualisation (NFV); Management and Orchestration}},
  institution  = {European Telecommunications Standards Institute (ETSI)},
  number       = {ETSI GS NFV-MAN 001 V1.1.1},
  year         = {2014},
  month        = dec
}

@techreport{etsi_mec003,
  author       = {{ETSI}},
  title        = {{Multi-access Edge Computing (MEC); Framework and Reference Architecture}},
  institution  = {European Telecommunications Standards Institute (ETSI)},
  number       = {ETSI GS MEC 003 V4.1.1},
  year         = {2025},
  month        = may
}

@techreport{3gpp23501,
  author       = {{3GPP}},
  title        = {{System Architecture for the 5G System (5GS)}},
  institution  = {3rd Generation Partnership Project (3GPP)},
  number       = {TS 23.501},
  year         = {2022}
}

@inproceedings{bonomi2012fog,
  author    = {Bonomi, Flavio and Milito, Rodolfo and Natarajan, Preethi and Zhu, Jiang},
  title     = {Fog Computing and Its Role in the Internet of Things},
  booktitle = {Proceedings of the First Edition of the MCC Workshop on Mobile Cloud Computing (MCC)},
  year      = {2012}
}

@article{bhamare2016sfc,
  author  = {Bhamare, Deval and Jain, Raj and Samaka, Mohammed and Erbad, Aiman},
  title   = {A Survey on Service Function Chaining},
  journal = {Journal of Network and Computer Applications},
  volume  = {75},
  pages   = {138--155},
  year    = {2016},
  month   = nov,
  doi     = {10.1016/j.jnca.2016.09.001}
}

@article{laghrissi2019vnfplacement,
  author  = {Laghrissi, Abdelquoddouss and Taleb, Tarik},
  title   = {A Survey on the Placement of Virtual Resources and Virtual Network Functions},
  journal = {IEEE Communications Surveys \& Tutorials},
  volume  = {21},
  number  = {2},
  pages   = {1409--1434},
  year    = {2019},
  doi     = {10.1109/COMST.2018.2884835}
}

@article{santos2020rlsfc,
  author  = {Santos, Guto Leoni and Lynn, Theo and Kelner, Judith and Endo, Patricia Takako},
  title   = {The Greatest Teacher, Failure is: Using Reinforcement Learning for SFC Placement Based on Availability and Energy Consumption},
  journal = {arXiv preprint},
  year    = {2020},
  note    = {arXiv:2010.05711}
}

@INPROCEEDINGS{ourconferencepaper,
  author={V\'azquez-Rodr\'{\i}guez, \'Alvaro and Fern\'andez-Veiga, Manuel and Giraldo-Rodr\'{\i}guez, Carlos},
  booktitle={2025 International Conference on Modeling, Analysis and Simulation of Wireless and Mobile Systems (MSWiM)}, 
  title={Diffusion-Based Solver for CNF Placement on the Cloud-Continuum}, 
  year={2025},
  volume={},
  number={},
  pages={476-482},
  doi={10.1109/MSWiM67937.2025.11309048}}





\end{document}